\documentclass[10pt,twocolumn,a4paper]{article}

\usepackage[final]{cvww}

\usepackage[pagebackref,breaklinks,colorlinks,allcolors=cvwwblue]{hyperref}

\usepackage[toc,acronym,style=long3col]{glossaries} 
\newacronym{LiDAR}{LiDAR}{Light Detection And Ranging}
\newacronym{ADAS}{ADAS}{Advanced Driver Assistance Systems}
\newacronym{AD}{AD}{Automated Driving}
\newacronym{DNNs}{DNNs}{Deep Neural Networks}
\newacronym{ZOD}{ZOD}{Zenseact Open Dataset}

\newacronym{FMCW}{FMCW}{Frequency-Modulated Continuous Wave}

\newacronym{KITTI}{KITTI}{Karlsruhe Institute of Technology and Toyota Technological Institute}
\newacronym{WOD}{WOD}{Waymo Open Dataset}
\newacronym{TP}{TP}{True Positive}
\newacronym{FP}{FP}{False Positive}
\newacronym{DOF}{DOF}{Degrees of Freedom}

\newacronym{ROS}{ROS}{Random Object Scaling}
\newacronym{OT}{OT}{Output Transform}
\newacronym{SN}{SN}{Statistical Normalization}
\newacronym{RBRS}{RBRS}{Random Beam Resampling}
\newacronym{UBRS}{UBRS}{Uniform Beam Resampling}

\newacronym{NLP}{NLP}{Natural Language Processing}

\newacronym{DNN}{DNN}{Deep Neural Networks}
\newacronym{GPU}{GPU}{Graphics Processing Unit}

\newacronym{DGC}{DGC}{Domain Gap Closed}
\newacronym{CNN}{CNN}{Convolutional Neural Network}
\newacronym{RNN}{RNN}{Recurrent Neural Networks}
\newacronym{mAP}{mAP}{Mean Average Precision}
\newacronym{AP}{AP}{Average Precision}
\newacronym{IOU}{IOU}{Intersection over Union}
\newacronym{ICG}{ICG}{Institute of Computer Graphics and Vision}
\newacronym{VFE}{VFE}{Voxel-Feature Encoding}
\newacronym{FCN}{FCN}{Fully Connected Network}
\newacronym{FPS}{FPS}{Farthest Point Sampling}
\newacronym{MSM}{MSM}{Masked Signal Modeling}
\newacronym{MAE}{MAE}{Masked Auto Encoder}
\newacronym{BEV}{BEV}{Bird-Eye-View}
\newacronym{SSL}{SSL}{Self-Supervised Learning}
\newacronym{RPN}{RPN}{Region Proposal Network}
\newacronym{NMS}{NMS}{Non Maximum Suppression}
\newacronym{MSA}{MSA}{Multi-Head Self-Attention}
\newacronym{FOV}{FOV}{Field of View}

\newacronym{ROI}{ROI}{Region of Interest}

\usepackage{booktabs}
\usepackage{multirow}
\usepackage{pgfplots}
\pgfplotsset{compat=1.9}
\usepackage{subcaption}
\usetikzlibrary{patterns}

\usepackage[edges]{forest}
\usepackage[normalem]{ulem}
\useunder{\uline}{\ul}{}
\def\paperID{12} 

\title{An Investigation of Beam Density on LiDAR Object Detection Performance}

\author{
Christoph Griesbacher$^{1}$\and 
Christian Fruhwirth-Reisinger$^{1,2}$\and\\
{$^{1}$Institute of Visual Computing, TU Graz}\\
{$^{2}$Christian Doppler Laboratory for Embedded Machine Learning}\\
{\tt\small \{griesbacher, reisinger\}@tugraz.at}
}
\begin{document}
\maketitle


\begin{abstract}
Accurate 3D object detection is a critical component of autonomous driving, enabling vehicles to perceive their surroundings with precision and make informed decisions. LiDAR sensors, widely used for their ability to provide detailed 3D measurements, are key to achieving this capability. However, variations between training and inference data can cause significant performance drops when object detection models are employed in different sensor settings. One critical factor is beam density, as inference on sparse, cost-effective LiDAR sensors is often preferred in real-world applications. Despite previous work addressing the beam-density-induced domain gap, substantial knowledge gaps remain, particularly concerning dense 128-beam sensors in cross-domain scenarios.

To gain better understanding of the impact of beam density on domain gaps, we conduct a comprehensive investigation that includes an evaluation of different object detection architectures. Our architecture evaluation reveals that combining voxel- and point-based approaches yields superior cross-domain performance by leveraging the strengths of both representations. Building on these findings, we analyze beam-density-induced domain gaps and argue that these domain gaps must be evaluated in conjunction with other domain shifts. Contrary to conventional beliefs, our experiments reveal that detectors benefit from training on denser data and exhibit robustness to beam density variations during inference.
\end{abstract}

\section{Introduction}
\label{sec:intro}

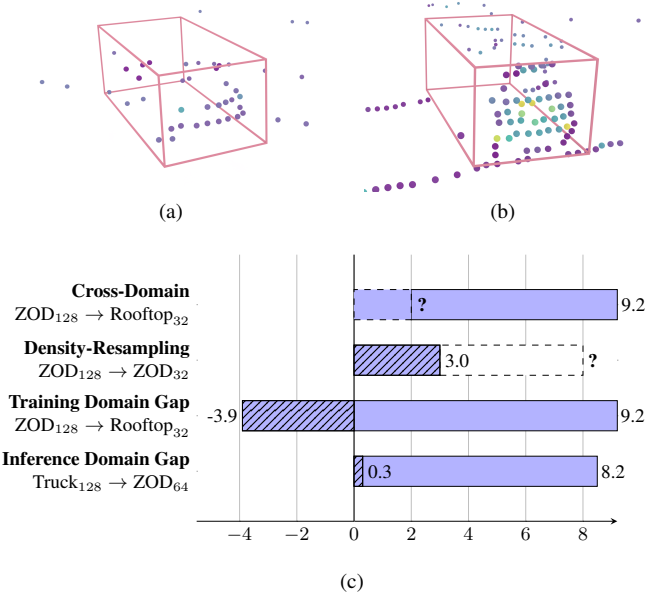
\begin{figure}[ht]
  \centering
  \begin{subfigure}{0.45\linewidth}
         \includegraphics[trim={10cm 10cm 60cm 20cm},clip,width=\textwidth]{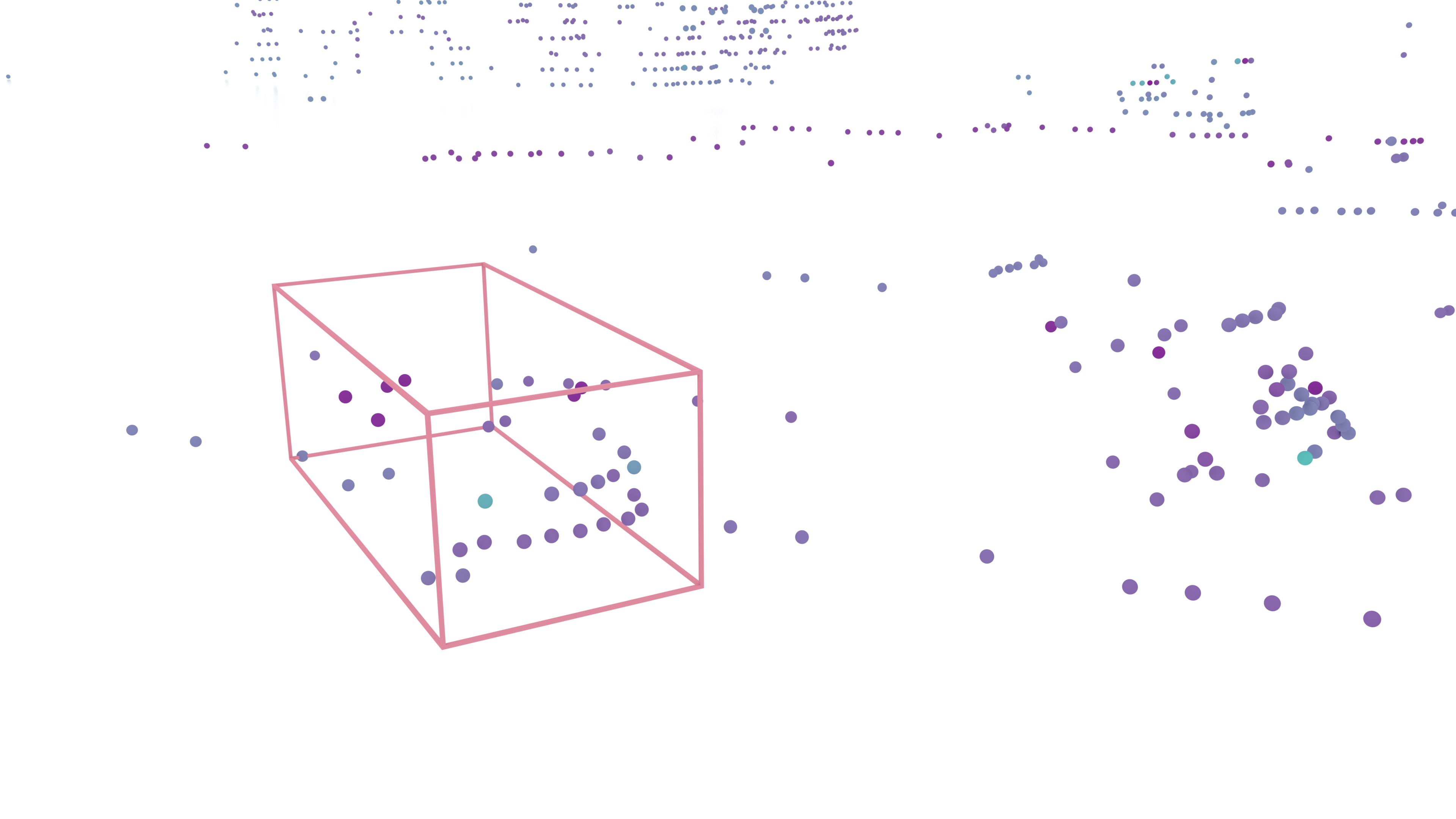}
    \caption{}
    \label{fig: sparse car}
  \end{subfigure}
  \hfill
  \begin{subfigure}{0.45\linewidth}
         \includegraphics[trim={40cm 9cm 10cm 8cm},clip,width=\textwidth]{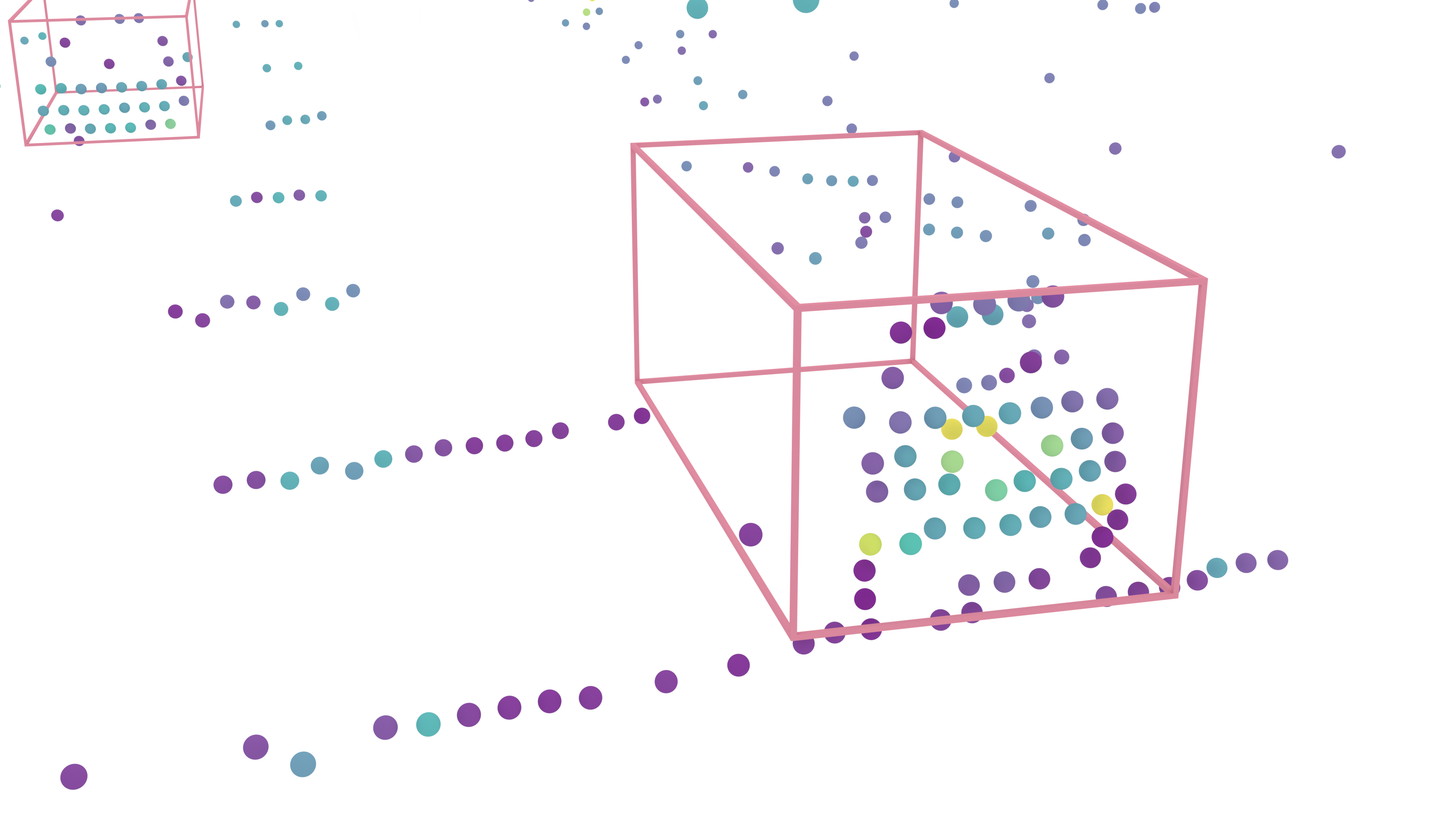}
    \caption{}
    \label{fig: dense car}
  \end{subfigure}
  \vspace{1em}
  \\
  \small
  \hspace{-5.5cm}
  \captionsetup[subfigure]{margin=-2.9cm}
    \begin{subfigure}{0.45\linewidth}
    
\pgfplotsset{
  custom labels/.style={
    nodes near coords,
    every node near coord/.append style={
      anchor=west, inner sep=1pt
    },
    visualization depends on={value \thisrow{label} \as \coordlabel},
    nodes near coords={\coordlabel}
  }
}

    \begin{tikzpicture}[scale=0.80] 
      \begin{axis}[
        xbar stacked,
        height=6cm,
        width=8.5cm,
        y axis line style = { opacity = 0 },
        xmin = -5.5,
        xmajorgrids,
        extra x ticks={0},
        extra x tick style={grid style={black}, xticklabel=\empty}, 
        stack negative=separate,
        enlarge y limits = 0.3,
        bar width = 0.5cm, 
        tickwidth         = 0pt,
        enlarge x limits = 0.02,
        ytick={0,1,2,3},
        yticklabels={
        {\textbf{Inference Domain Gap}\\Truck$_{128} \rightarrow \text{ZOD}_{64}$},
        \textbf{Training Domain Gap}\\ZOD$_{128} \rightarrow \text{Rooftop}_{32}$, 
        \textbf{Density-Resampling}\\ZOD$_{128} \rightarrow \text{ZOD}_{32}$, 
        \textbf{Cross-Domain}\\ZOD$_{128} \rightarrow \text{Rooftop}_{32}$, 
        },
        nodes near coords=\pgfmathsetmacro{\mystring}{{\mylst}[\coordindex]}\mystring,
        nodes near coords align={east},
        axis x line = bottom,
        yticklabel style={align=right}]
         legend style={
            area legend,
            at={(0.5,-0.15)},
            anchor=north,
            legend columns=-1,
        }, 
    \addplot[
      fill=blue!30!white,
      draw=black, dashed,
      custom labels
    ] table[row sep=\\] {
      x   y   label \\
      0  0   {} \\     
      0  1   {} \\
      0  2   {} \\
      2  3   {\hspace{0.45cm} \textbf{?}} \\
    };
    
    \addplot[
      fill=blue!30!white,
      postaction={pattern=north east lines},
      custom labels
    ] table[row sep=\\] {
      x   y   label \\
      0.3  0   {\hspace{0.12cm}0.3} \\     
      -3.9  1   {\hspace{-2.55cm}-3.9} \\
      3  2   {\hspace{0.75cm}3.0} \\
      0  3   {} \\
    };

    \addplot[
      fill=white,
      draw=black, dashed,
      custom labels
    ] table[row sep=\\] {
      x   y   label \\
      0  0   {} \\
      0  1   {} \\
      5.0  2   {\hspace{1.22cm}\textbf{?}} \\       
      0  3   {} \\
    };

    \addplot[
      fill=blue!30!white,
      custom labels
    ] table[row sep=\\] {
      x   y   label \\
      8.2  0   {\hspace{1.95cm}8.2} \\
      9.2  1   {\hspace{2.18cm}9.2} \\
      0  2   {} \\
      7.2  3   {\hspace{1.72cm}9.2} \\       
    };
    
      \end{axis}
    \end{tikzpicture}
    \caption{\hspace{-3cm}}
    \label{fig: domain gap assessment methods}
  \end{subfigure}

\caption{(a) Low-density and (b) high-density scan of vehicles at a similar distance. (c) Overall and beam-density-induced domain gap (in \% for IOU=0.4) measured by different methods. The Cross-Domain and Density-Resampling methods fail to assess either the beam-density-induced or overall domain gap, while the Training and Inference Domain Gaps provide a complete picture. \vspace{-0.4cm}}
\label{fig: teaser figure}
\end{figure}
Autonomous driving has been receiving increasing attention in recent years, as it has the potential to increase road safety, traffic efficiency, and reduce emissions. 
To enable the decision-making capabilities of Advanced Driver Assistance Systems (ADAS) or Automated Driving (AD) technologies, understanding the vehicle's immediate environment is crucial. Light Detection And Ranging (LiDAR) technology stands out as a particularly effective solution for this task through its ability to directly measure three-dimensional distances with high accuracy~\cite{mao_3d_2022}. 

LiDAR-based 3D object detection models have demonstrated impressive performance on established benchmarks~\cite{geiger_are_2012, caesar_nuscenes_2020, sun_scalability_2020, mao_one_2021, diaz-ruiz_ithaca365_2022}. However, their performance often drops significantly when applied across different datasets due to inherent differences between the source domain and the target domain. Typical examples are varying sensor configurations between source and target domain or adverse weather conditions covered by the target but not the source data. When these differences are substantial, the detection model struggles to generalize to the new domain, introducing a performance gap known as the \emph{domain gap}. This challenge is particularly critical in real-world applications, where a domain gap is almost inevitable due to the variability between the training dataset and the diverse conditions encountered in deployment.

The common way to mitigate domain gaps is the application of domain adaptation methods. Such methods are oftentimes tailored towards a specific domain difference, such as different LiDAR resolutions~\cite{hu_density-insensitive_2023, li_domain_2023, wei_lidar_2022} or varying object size distributions~\cite{wang_train_2020, yang_st3d_2021, malic_sailor_2023}. Thus, to successfully apply domain adaptation methods, the most significant domain shifts have to be identified first. A structured domain shift taxonomy is useful in this context, as it helps to categorize and systematically understand the specific shifts between domains, enabling the selection or design of a targeted adaptation technique.

Despite taxonomies of related work~\cite{eskandar_empirical_2024, manivasagam_towards_2023, yu_benchmarking_2023, dong_benchmarking_2023}, no study includes all the domain shifts between the datasets under investigation (described in \cref{chap: dataset intro}). Thus, we introduce a domain shift taxonomy in ~\cref{fig: domain shift taxonomy}. Keeping the goal of domain adaptation in mind, we distinguish between domain shifts that can effectively be addressed by domain adaptation methods and those that persist despite the application of domain adaptation. We call the former \emph{non-persistent} and the later \emph{persistent} domain shifts.

Motivated by the observation that the domain gap varies significantly when employing different detectors, we conduct an object detector architecture evaluation. While prior studies do not give particular attention to a thoughtful selection of an object detection model~\cite{richter_understanding_2022}, we aim to identify detectors that are inherently robust against domain changes. By minimizing the initial domain gap, the reliance on domain adaptation is minimized, ensuring that the domain adaptation efforts focus on the most challenging aspects of domain gap. We find that (1) voxel-based detectors robustly detect objects, but have difficulties at precisely localizing them and (2) point-based detectors excel at localizing objects in cross-domain settings. Our experiments suggest that optimal cross-domain detection performance is achieved by combining voxel- and point-based approaches in a two-staged detector.

A particularly important domain shift stems from the number of LiDAR beams (see~\cref{fig: teaser figure}). High-density LiDAR sensors produce detailed point clouds with a high number of points, easing the accurate estimation of object sizes and positions. Low-density LiDAR sensors, which are often more affordable and more commonly used in large-scale deployments, capture fewer points, leading to sparser point clouds and less reliable detection results. This difference in beam density creates a domain shift when models trained on high-density LiDAR data are applied to low-density data and vice versa.
To analyze the domain gap caused by varying beam densities, related studies utilize one of two approaches. The first approach~\cite{hu_density-insensitive_2023} involves multiple datasets employing LiDAR sensors with varying beam densities which are subsequently compared. The second approach~\cite{wei_lidar_2022, eskandar_empirical_2024, richter_understanding_2022} is based on downsampling a dense dataset to create sparser twin-dataset with varying beam density which are subsequently compared. 

This paper highlights the shortcomings of the existing methods. First, comparing the domain gap between two datasets does not guarantee that the observed domain gap actually stems from varying beam density or is caused by other domain shifts occurring between the investigated datasets. The second approach leads to ambiguous results because it analyzes the effect of beam density in isolation of other domain shifts. In real-world applications, the beam-density-induced domain gap is always accompanied by other effects influencing the domain gap. We show that the beam-density-caused domain gap has to be assessed \emph{in conjunction} with other domain shifts to accurately evaluate its impact in real-world applications. Our experiments suggest that (1) in contrast to the results of related studies~\cite{eskandar_empirical_2024, fang_lidar-cs_2024, richter_understanding_2022}, it is more beneficial to train object detectors on dense data, independent of the density of the target data and (2) concerning the inference domain gap, detectors are robust against a change of up to 64 beams (see~\cref{fig: domain gap assessment methods}).
\noindent
Our contributions can be summarized as follows:
\begin{itemize}
    \item We introduce a domain shift taxonomy based on macro-, sensor-, and object-level domain shifts and distinguish between persistent and non-persistent domain shifts.
    \item We conduct a detector architecture evaluation where we compare different detectors by their inherent domain adaptation abilities.
    \item We investigate the domain gap induced by varying beam densities including 128-beam sensors on real-world datasets with consideration of other domain shifts.
\end{itemize}

\section{Related Work}
\label{sec:related work}

\noindent
\textbf{Object Detection:}
In LiDAR-based 3D object detection, architectural choices heavily influence detection performance. Voxel-based methods~\cite{zhou_voxelnet_2018, yan_second_2018}, discretize LiDAR points into 3D grids, allowing for efficient feature extraction through sparse convolutions. Pillar-based methods~\cite{lang_pointpillars_2019, shi_pillarnet_2022, li_pillarnext_2023, fan_embracing_2022} convert the point cloud into a 2D BEV-image, sacrificing height information for computational efficiency. Operating directly on the raw points, point-based approaches~\cite{qi_pointnet_2017-1, pan_3d_2021, zhang_not_2022} retain spatial details without quantization. Recent transformer-based models~\cite{wang_dsvt_2023, zhou_octr_2023, zhou2024lidarformer}, provide an alternative to CNN-based models~\cite{zhang_hednet_2023, zhang_safdnet_2024, chen_largekernel3d_2023}, capturing interactions across larger spatial regions. Detection heads in object detection are either anchor-based~\cite{liu_ssd_2016}, relying on predefined anchor sizes, or anchor-free~\cite{yin_centerpoint_2021}, which directly predict object centers to generate bounding boxes. Two-staged detectors~\cite{shi_pointrcnn_2019, shi_pv-rcnn_2022, shi_pv-rcnn_2022} split the detection into a proposal and refinement stage, often improving accuracy over single-stage detectors but at a higher computational cost. 

Concurrent to our work, Eskandar \emph{et al.}~\cite{eskandar_empirical_2024} empirically test the impact of fundamental architecture choices. However, they chose different detectors to represent each architectural choice. While Eskandar \emph{et al.} choose Point-RCNN~\cite{shi_pointrcnn_2019}, VoTr~\cite{mao_voxel_2021} and PV-RCNN~\cite{shi_pv-rcnn_2020}, we select the faster or better performing object detectors IA-SSD~\cite{zhang_not_2022}, DSVT~\cite{wang_dsvt_2023} and PV-RCNN++~\cite{shi_pv-rcnn_2022} for the point-based, Transformer-based and two-staged architectures.

\noindent
\textbf{Domain Gap Analysis:}
Recent works have extensively studied how specific domain shifts contribute to the overall domain gap. Wang~\emph{et al.}~\cite{wang_train_2020} analyzed the impact of geographical variations, concluding that differences in object size distribution can significantly affect detection performance. Another well-studied factor is weather~\cite{hegde_adverse_2023, dreissig_survey_2023}: while LiDAR sensors are less susceptible to adverse weather than cameras, conditions such as snow~\cite{kurup_dsor_2021}, rain~\cite{xu_spg_2021}, or fog~\cite{kim_empirical_2023} still impair object detection. Concerning sensor-level domain shifts, Hu~\emph{et al.}~\cite{hu_investigating_2022} and Fang~\emph{et al.}~\cite{fang_lidar-cs_2024} investigate the impact of different LiDAR mounting positions. There are also some recent works investigating the effect of varying beam densities~\cite{richter_understanding_2022, fang_lidar-cs_2024}. Richter \emph{et al.}~\cite{richter_understanding_2022} perform a real-world study comparing a 32-beam and 64-beam LiDAR sensors utilizing a specially designed dataset, isolating the beam-density-induced domain gap. However, they do not analyze beam density in conjunction with other domain shifts such as geographic location or object size. Fang~\emph{et al.}~\cite{fang_lidar-cs_2024} perform a systematic study regarding beam density on a simulated dataset. However, they did not test the transferability of their findings to real-world datasets. 

\noindent
\textbf{Domain Adaptation:}
Domain adaptation methods aim to improve object detection performance across different datasets, addressing the challenges introduced by domain shifts. Broadly, domain adaptation approaches fall into one of three categories: domain alignment, feature alignment, and self-training.

The domain alignment methods SN~\cite{wang_train_2020}, OT~\cite{wang_train_2020}  and SAILOR~\cite{malic_sailor_2023} excel at handling object size discrepancies by rescaling ground truth bounding boxes during training or inference. For beam density shifts, methods like DTS~\cite{hu_density-insensitive_2023}, PDDA~\cite{li_domain_2023} and LiDAR-CS~\cite{fang_lidar-cs_2024} employ resampling methods to align point cloud densities. ReSimAD~\cite{zhang_resimad_2024} aligns more complex LiDAR sensor characteristics by reconstructing target scenes and rendering source-like point clouds. Feature alignment methods~\cite{wozniak_uada3d_2024, zhang_srdan_2021, wei_lidar_2022, luo_unsupervised_2021}, another approach, perform domain adaptation by alignment in feature space instead of aligning the point clouds directly. In self-training~\cite{saltori_sf-uda_2020, peng_cl3d_2023, yang_st3d_2021, fruhwirth-reisinger_fast3d_2021, chen_revisiting_2023, yang_st3d_2022}, iterative refinement of pseudo-labels is used to gradually adapt the detector to the target domain.

While these domain adaptation methods effectively reduce the occurring domain gaps, they pay little attention to the underlying object detector. We show that a thoughtful selection of the object detector architecture can already close a portion of the domain gap, which reduces the reliance on domain adaptation methods and shifts the focus to more complex domain shifts which cannot be mitigated through architecture alone.

\section{Preliminary Analysis}
\label{sec:preliminaries}
Our preliminary analysis lays the groundwork for this study by addressing three aspects. First, we introduce the non-public datasets involved in this study and detail their unique properties. Second, we establish a domain shift taxonomy, allowing us to systematically assess domain differences. Third, we conduct a detector architecture evaluation to identify models that are inherently robust to domain shifts, providing a foundation for effective domain adaptation.

\subsection{Dataset Introduction}
\label{chap: dataset intro}

\begin{figure*}[t]
  \centering
  \begin{subfigure}{0.3\linewidth}
         \includegraphics[width=\textwidth]{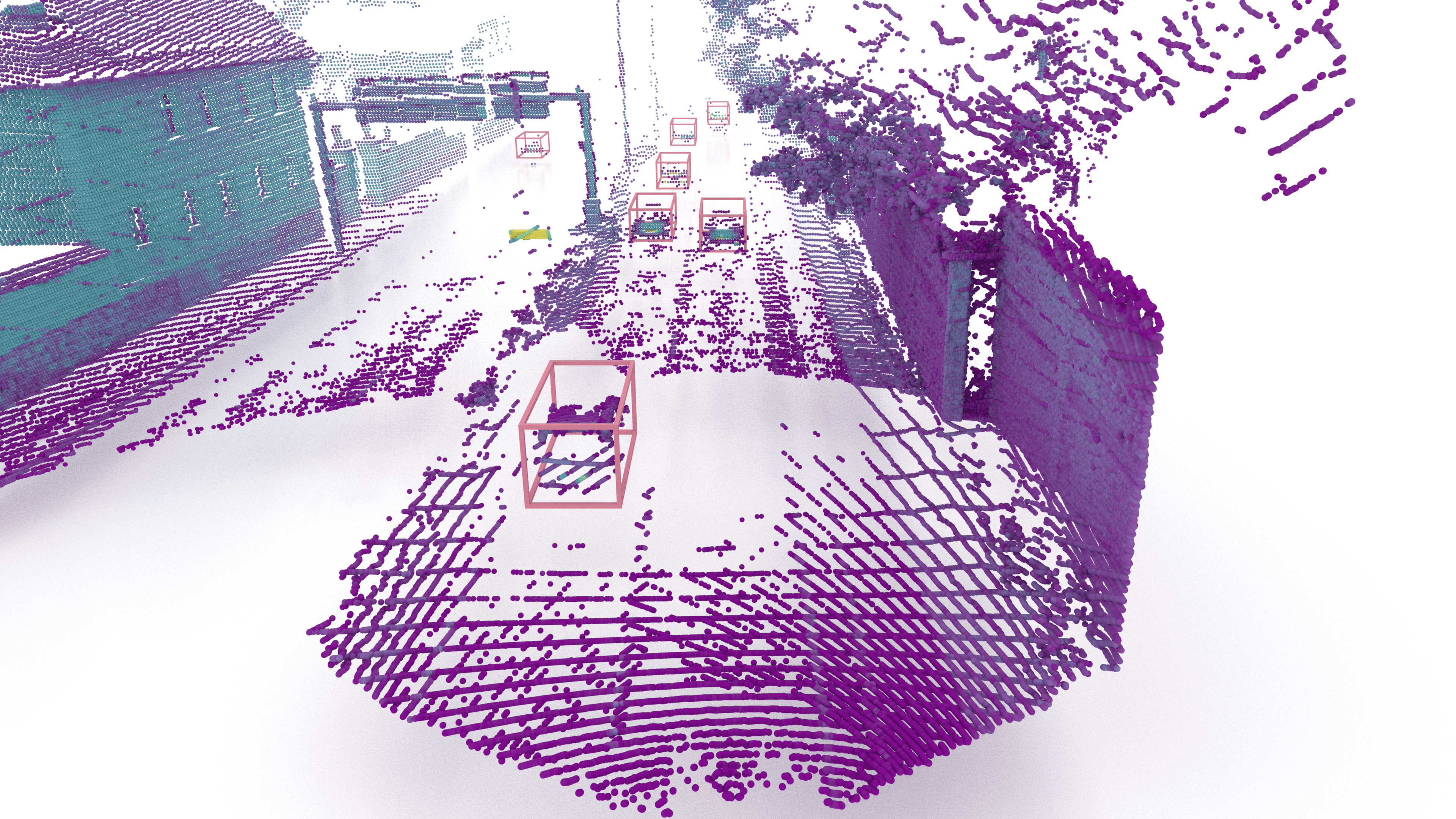}
    \caption{}
  \end{subfigure}
  \hfill
  \begin{subfigure}{0.3\linewidth}
    \includegraphics[width=\textwidth]{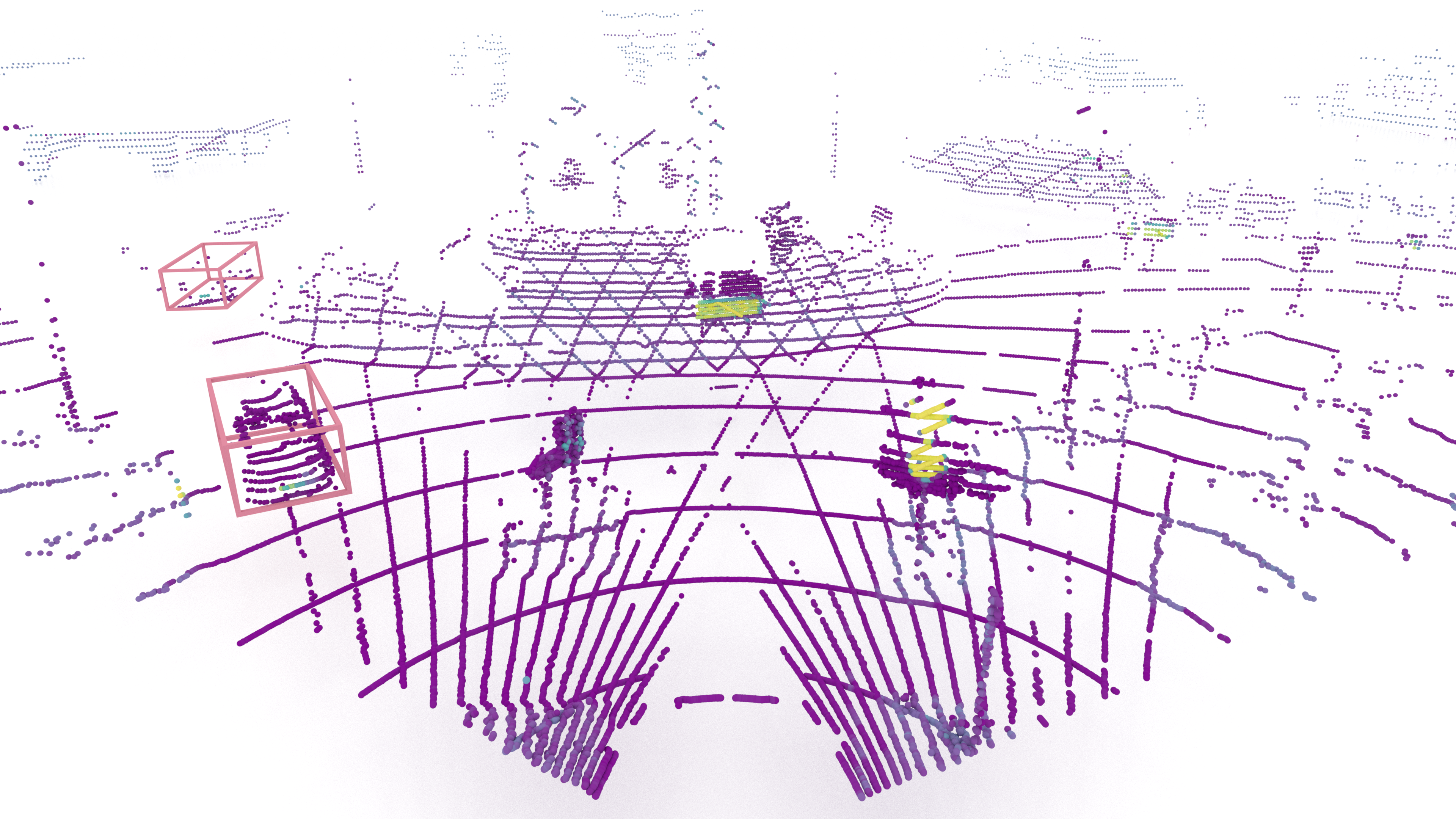}
    \caption{}
  \end{subfigure}
\hfill
  \begin{subfigure}{0.3\linewidth}
         \includegraphics[width=\textwidth]{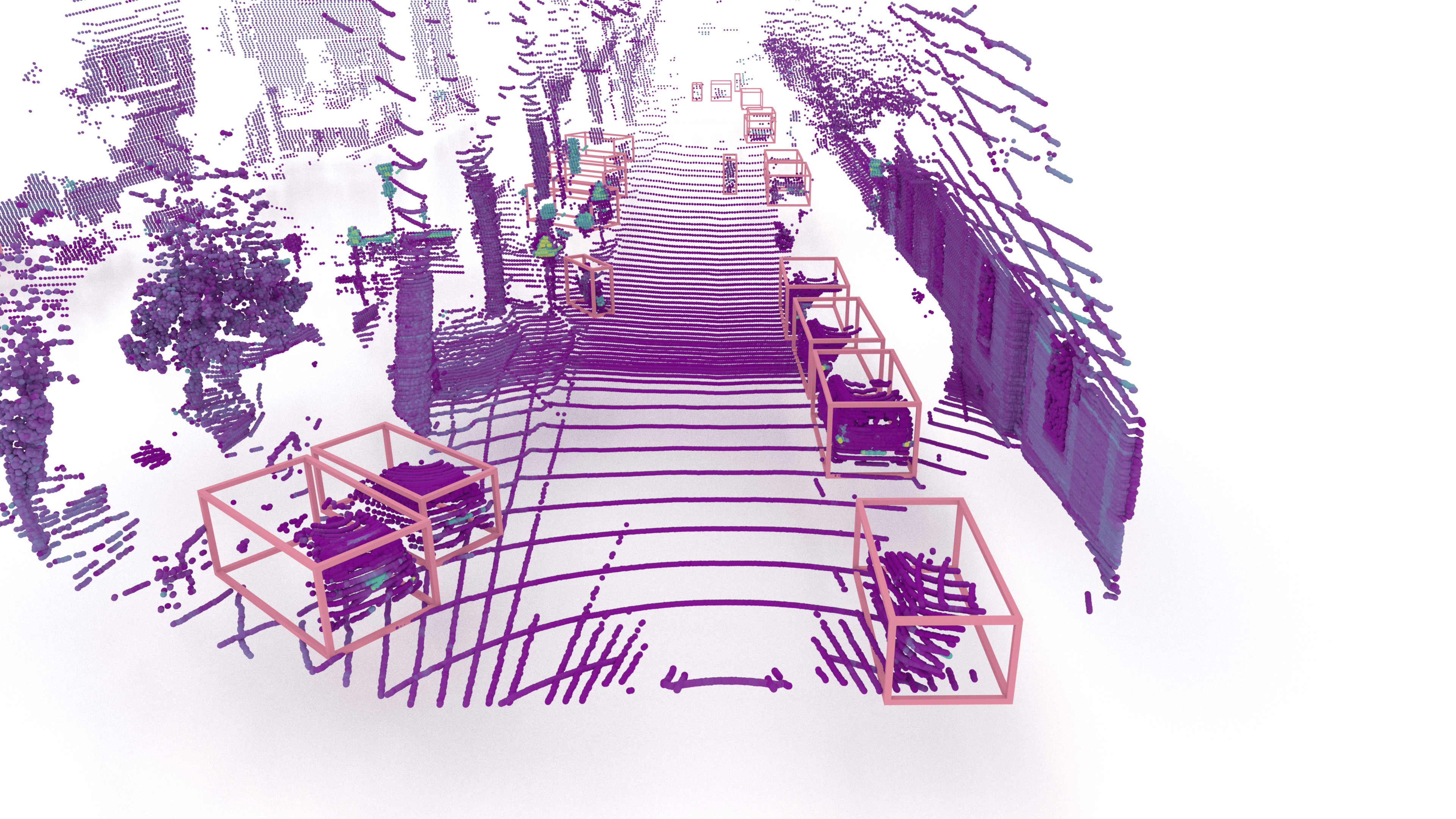}
    \caption{}
  \end{subfigure}
\caption[Dataset comparison]{Comparison of the (a) Truck, (b) Rooftop and (c) Zenseact datasets. Differences in LiDAR beam density are clearly visible. Ground-truth objects are marked by red bounding boxes. \vspace{-0.4cm}}
\label{fig: dataset comparison}
\end{figure*}

\begin{table}[t]
\setlength{\tabcolsep}{3pt} 
\centering
\begin{tabular}{@{}lccc@{}}
\toprule
                                               & \textbf{Truck}                                                 & \textbf{Rooftop}                                                 & \textbf{ZOD}                                                         \\ \midrule
\multicolumn{1}{l|}{Locations}                 & Germany                                                   & Germany                                                     & \begin{tabular}[c]{@{}c@{}} 15 European \\ Countries \end{tabular}\\
\multicolumn{1}{l|}{Ann. frames}               & 40k                                                       & 7.5k                                                        & 100k                                                                          \\
\multicolumn{1}{l|}{Sequences}                 & 2036                                                      & 251                                                         & 43468                                                                         \\ \midrule
\multicolumn{1}{l|}{Top LiDAR}                 & \begin{tabular}[c]{@{}c@{}}OS2 \\ (128-beam)\end{tabular} & \begin{tabular}[c]{@{}c@{}}VLP 32c\\ (32-beam)\end{tabular} & \begin{tabular}[c]{@{}c@{}}VLS 128\\  (128-beam)\end{tabular}                 \\
\multicolumn{1}{l|}{Mounting height}           & 3.41m                                                     & 1.78m                                                       & 2.01m                                                                         \\
\multicolumn{1}{l|}{Side LiDARs}               & 64-beam                                                   & 16-beam                                                     & 16-beam                                                             \\
\multicolumn{1}{l|}{Front LiDAR}               & 32-beam                                                   & -                                                           & -                                                                             \\
\multicolumn{1}{l|}{Avg. pts per frame}     & 178.4k                                                    & 71.5k                                                       & 254k                                                                          \\
\multicolumn{1}{l|}{Points per beam}           & 2048                                                      & 1800                                                        & 3270   \\
\multicolumn{1}{l|}{Horizontal res.}           & 0.18°                                                      & 0.2°                                                        & 0.11°       \\ \midrule
\multicolumn{1}{l|}{License}                   & private                & private                                                                                          & CC BY-SA \\ \bottomrule
\end{tabular}
\caption{Dataset overview. \vspace{-0.4cm}}
\label{tab: dataset comparison}
\setlength{\tabcolsep}{6pt} 
\end{table}

In this paper we leverage three datasets (see~\cref{tab: dataset comparison} and~\cref{fig: dataset comparison}) for training and evaluation. The Rooftop and Truck datasets are private while the remaining Zenseact Open Dataset (ZOD)~\cite{alibeigi_zenseact_2023} is open-source. All datasets are specifically designed for autonomous driving applications and feature frame-wise LiDAR data. Concerning the dataset size, the Rooftop dataset is the smallest, with about 7.5k annotated frames, while the Truck and Zenseact Open datasets are substantially larger with 40k and 100k annotated frames. While the Rooftop and Truck datasets were both recorded in Germany, the ZOD contains data from 15 different European countries. Another substantial difference concerns the organization of frames. The Rooftop and Truck datasets are structured in sequences of 20 or 30 frames per sequence, while the ZOD consists of single frames, where, on average, only two frames belong to the same sequence. Regarding the sensor setup, the main LiDAR also differs between each dataset. The Zenseact and Truck datasets employ a dense 128-beam LiDAR, each from a different manufacturer, while the Rooftop dataset employs a sparse 32-beam main LiDAR. We conduct a detailed analysis of the differences between the datasets in the subsequent~\cref{chap: domain shift taxonomy}. 

Some inherent differences between the datasets can be eliminated by dataset alignment. The size difference can be aligned rather easily by randomly subsampling of the larger datasets to match the size of the smallest dataset. We identified three major dataset alignment measures to address the frame content: coordinate, range and label-space alignment.\\  
\textbf{Coordinate Alignment:} Datasets often differ in coordinate systems, leading to potential mismatches between LiDAR points and object labels. To address this, we align all data points and labels with the commonly used sensor coordinate system with a forward-pointing x-axis and an upward-pointing z-axis. Since sensors are mounted at different heights, we standardize the origin by aligning it with the ground plane.\\
\textbf{Range Alignment:} Standardizing the Field of View (FOV) across datasets allows the detector to learn consistent object regions. We define a forward detection range of 123.2 meters to support high-speed safety applications and limit the horizontal FOV to 120° to match the ZOD’s labeled region. We mark objects truncated by the cropped FOV as “ignore” during training and evaluation, which prevents the generation of a loss from these objects.  \\
\textbf{Label-Space Alignment:} Inherent label-space differences of the ground-truth annotations between datasets necessitate a mapping to standardize object classes. We categorize objects into four primary classes: \emph{Vehicle}, \emph{Truck}, \emph{Single-track}, and \emph{Pedestrian}. Single-track vehicles are composed of bicycle and motorcycles. Larger vehicles such as vans, trucks and trailers fall under the \emph{Truck} class. To handle varying labeling conventions concerning single-tracked vehicles and their riders, we merge their bounding boxes encompassing both as a single object. A detailed mapping of the label-spaces between the three datasets can be found in the Supplementary.

\subsection{Domain Shift Taxonomy}
\label{chap: domain shift taxonomy}

\setlength{\tabcolsep}{3pt} 
\begin{table}[t]
\centering
\begin{tabular}{@{}lcc@{}}
\toprule
 & Persistent & Non-persistent \\ \midrule
\multirow{3}{*}{\textbf{\begin{tabular}[c]{@{}l@{}}Macro-\\level \end{tabular}}}
& Collection Area Type  &   Object Size Statistics \\
&  Geographical Location &   Weather Conditions \\
 & Frame Selection &\\ \midrule
 \multirow{4}{*}{\textbf{\begin{tabular}[c]{@{}l@{}}Sensor-\\level \end{tabular}}}  
 & Sensor Setup & Beam Density \\
 & Intensity Value & Horizontal Resolution \\
 & Rate of Rotation &  Field of View\\ 
 & Alignment Error & \\\midrule
 \multirow{3}{*}{\textbf{\begin{tabular}[c]{@{}l@{}}Object-\\level \end{tabular}}}  
 & Labeling Quality& Label Space Definition \\
 & & Labeling Zone \\
 & & Object Definition\\\midrule 
\end{tabular}
\caption{Domain shift taxonomy. We differentiate between persistent and non-persistent to highlight domain shifts that can effectively be addressed by either readily available domain adaptation methods or dataset alignment measures. \vspace{-0.4cm}}
\label{fig: domain shift taxonomy}
\setlength{\tabcolsep}{6pt} 
\end{table}
We propose a domain shift taxonomy which allows for a detailed and systematic investigation of possibly occurring domain shifts between aligned datasets. We distinguish between three main categories. Sensor-level domain shifts are directly caused by the mode of collection, while Object-level domain shifts concern the object definition and labeling. The remaining macro-level domain shifts are mainly caused by differences in dataset content. Keeping the final goal of domain adaptation in mind, we additionally differentiate between domain shifts that can effectively be reduced by domain adaptation methods, the \emph{non-persistent domain shifts}, and those that persist despite domain adaptation methods, which we refer to as \emph{persistent domain shifts}.

Concerning the \emph{persistent domain shifts}, we notice a few macro-level differences. While the ZOD features a geographically diverse set of recording locations, the Rooftop and Truck datasets were exclusively recorded in Germany. Also, the types of areas differ between datasets: the ZOD features substantially more \emph{City} frames compared to the remaining two datasets. Finally, we find differences that likely originate from the frame selection process for each dataset. We notice that there is a significantly lower number of overall objects in the Rooftop dataset compared to the Truck and Zenseact datasets. Especially the \emph{Pedestrian} and \emph{Cyclist} classes are significantly underrepresented such that the missing diversity of classes would dominate the domain gap. Thus, we resort to mainly perform dataset-wise comparisons between the \emph{Vehicle} classes.

There are also significant differences on a sensor-level. The Truck dataset has a unique sensor setup as the sensors are mounted considerably higher compared to the other two datasets. The high mounting position has the consequence of a large blind spot right in front of the ego vehicle. The installation of an additional forward-facing LiDAR addresses this issue, resulting in a four-sensor setup.

In terms of object-level differences, we find disparities between the datasets caused by deficient labeling. More specifically, we notice missing ground truth labels for the Rooftop dataset, especially for distant objects that are hit by less LiDAR points. The implications are a noisy supervision signal for training and a distorted evaluation result as predominantly hard-to-detect objects are missing. The ZOD suffers from a similar problem, but hereby, the missing labels are caused by the labeling procedure. ZOD's labeling is based on the camera images. Slight height differences between the camera and LiDAR sensors cause objects to be occluded for the camera while visible for the LiDAR, resulting in missing labels. 

We also identify many \emph{non-persistent domain shifts}. In contrast to the previous class of domain shifts, the non-persistent ones can effectively be reduced or even eliminated by domain adaptation methods. Most prominently, the datasets employ LiDAR sensors with a differing number of beams as well as varying beam patterns. Furthermore, the ZOD is more diverse in terms of captured weather conditions as it also features adverse weather conditions such as fog or snow. We also notice differences in terms of object sizes. As the ZOD contains frames recorded in multiple different countries, the intra-dataset object size variability is higher. 

Throughout our analysis, we find numerous domain shifts between the datasets. Most of the identified shifts cannot be isolated, making it infeasible to estimate the impact of individual domain shifts on the overall domain gap by a simple comparison between datasets. 
We provide detailed statistics and domain shift examples in the Supplementary Material.
\subsection{Detector Architecture Evaluation}
\begin{table}[t]
\centering
\begin{tabular}{@{}lcccr@{}}
 & \multicolumn{1}{l}{} & \multicolumn{1}{l}{} & \multicolumn{1}{l}{} & \multicolumn{1}{l}{}  \\ \midrule
\textbf{Detector} & \multicolumn{2}{c}{\textbf{\begin{tabular}[c]{@{}c@{}}Backbone\\ 
Architecture\end{tabular}}} & \textbf{\begin{tabular}[c]{@{}c@{}}Detection\\ Head\end{tabular}} & \textbf{Stages} \\ \midrule
SECOND~\cite{yan_second_2018} & Voxel & CNN & Anchor & Single  \\
PointPillar~\cite{lang_pointpillars_2019} & Pillar & CNN & Anchor & Single  \\
IA-SSD~\cite{zhang_not_2022} & Point & CNN & Point & Single  \\
CenterPoint~\cite{yin_centerpoint_2021}& Voxel & CNN & Center & Single  \\
PVRCNN++~\cite{shi_pv-rcnn_2022} & \begin{tabular}[c]{@{}c@{}}Point- \\ Voxel\end{tabular} & CNN & \begin{tabular}[c]{@{}c@{}}Center/\\ Point\end{tabular} & Two  \\
DSVT~\cite{wang_dsvt_2023} & Pillar & \begin{tabular}[c]{@{}c@{}}Trans-\\ former\end{tabular} & Center & Single  \\ \bottomrule
\end{tabular}
\caption{List of 3D object detection methods and their architectural properties. \vspace{-0.4cm}}
\label{tab: detector comparison}
\end{table}

We identify six key differences among commonly used object detection architectures and select one object detector representative of each difference. This approach allows us to assess the impact of each architectural choice. An overview of the selected object detectors is given in~\cref{tab: detector comparison}. In terms of data representation, we choose SECOND~\cite{yan_second_2018} to represent voxel-based architectures, PointPillars~\cite{lang_pointpillars_2019} for the pillar-based representation, and IA-SSD~\cite{zhang_not_2022} to represent the class of point-based object detectors. Furthermore, we select CenterPoint~\cite{yin_centerpoint_2021} to assess the effect of center-based detection heads. To reason about the effectiveness of two-staged methods, we employ the point-voxel-based detector PV-RCNN++~\cite{shi_pv-rcnn_2022}. This detector uses a SECOND-like first-stage to extract bounding box proposals and a point-feature-based second-stage to refine the proposals for the final bounding box estimation. Lastly, we test the impact of different feature extractor architectures. As Transformer-based architectures have recently established themselves in the field of 3D object detection~\cite{wang_dsvt_2023}, we test their performance in comparison to the well-established sparse-convolution-based architectures.

\section{Approach}
\label{sec:approach}
To evaluate the impact of the beam density on the cross-domain performance, we first select a detector architecture that demonstrates robustness across domains. In our initial experiments, we simply evaluate the trained detectors across domains and group the detection results according to the domain shifts of interest. Results are reported in both high- and low-IOU settings to differentiate between localization and detection errors. For our analysis, we primarily focus on detection errors, which are assessed using low-IOU experiments, as localization errors can usually be mitigated through domain adaptation methods targeting object sizes~\cite{wang_train_2020, malic_sailor_2023}.
This cross-domain comparison operates under the assumption that the datasets are sufficiently similar to enable meaningful conclusions. However, as elaborated in \cref{chap: domain shift taxonomy}, this assumption rarely holds due to persistent domain gaps arising from differences in sensor setups, environmental conditions, and other factors.

In our second set of experiments, we analyze the impact of varying beam densities by isolating it from other domain shifts. Following the approach of \cite{eskandar_empirical_2024, wei_lidar_2022}, we generate beam-wise downsampled versions of one dataset. A detector is trained on each version and subsequently evaluated on the other versions of the same dataset. This approach allows us to focus on the impact of beam density, independent of other domain-specific properties.
However, in real-world applications, a varying beam density is usually just one of many domain shifts occurring at test time. In such cases, other kinds of domain shifts may completely dominate the domain gap, rendering the effect of beam density negligible. On the contrary, it could also be the case that due to the cross-domain application, other more reliable features are missing, resulting in an increased domain gap caused by varying beam density. Focusing on a single domain shift in isolation fails to capture these complex interactions. Thus, more sophisticated experiments with the goal of capturing the domain gap by a certain domain shift \emph{in conjunction} with other domain shifts is necessary.
 
To address the limitations of isolated domain shift analysis, we propose an experimental setup designed to account for interactions between beam density and other domain shifts. As in prior experiments, we utilize sparsified versions of datasets to analyze the impact of beam density, but this time in conjunction with other datasets. Our setup divides the domain gap into two components: the \emph{training domain gap}, caused by differences in beam density during training, and the \emph{inference domain gap}, caused by variations during evaluation, as described by Richter \emph{et al.}~\cite{richter_understanding_2022}. To measure the training domain gap, we downsample the training dataset to create multiple versions, each representing a specific beam density level. By matching or mismatching the beam density with the evaluation dataset, we isolate the effects of beam density variation during training. Similarly, the inference domain gap can be attained by varying the beam density of the evaluation datasets while keeping the training datasets unchanged. These controlled experiments allow us to isolate the specific effects of beam density in a cross-domain setting. We note that in real-world applications, it is typically infeasible to disentangle the training and inference domain gaps, underscoring the relevance of these controlled experiments.

\section{Experiments}
\label{sec:experiments}

To demonstrate the effectiveness of our assessment approach, we conduct experiments on the two private datasets Rooftop and Truck and the public Zenseact Open Dataset~\cite{alibeigi_zenseact_2023}. We first present our results for the object detector architecture evaluation, based on which we then assess the cross-domain and density-resampling domain gaps. Finally, we compare our training and inference domain gaps to the previously determined domain gaps. Detailed experiment results and additional information about implementation and model training can be found in the Supplementary Material. 
\subsection{Evaluation Metrics}
We use the Intersection-over-Union (IOU)-based metric \emph{3D average precision} $\mathrm{AP}_{S\rightarrow T}$ to assess the detection performance of an object detection model trained on the source domain $\mathbb{D}^S$ when evaluated on the target domain $\mathbb{D}^T$. By lowering the IOU threshold, we can additionally disentangle the \emph{detection error} from the \emph{localization error}. Thereby, localization errors are caused by objects that are detected but not localized accurately enough to be considered true positives, whereas detection errors represent entirely missed or wrongly classified objects. Wang \emph{et al.}~\cite{wang_train_2020} demonstrated that the 3D average precision significantly increases when the IOU threshold is reduced from the commonly used threshold of 0.7 (70\%) to approximately 0.4 (40\%). At this threshold, the domain gap primarily reflects detection errors, which are of greater practical significance than localization errors. Localization errors can often be mitigated using domain adaptation methods such as ROS~\cite{yang_st3d_2021}, SN, or OT~\cite{wang_train_2020}. Therefore, in our evaluation, we primarily focus on a reduced IOU threshold of 0.4 to better understand detection errors.

While the cross-domain performance is well suited for comparing different detectors, it does not adequately capture the generalization ability of a certain detector across domains, as it is influenced by the inherent difficulty of the target dataset. To address this limitation, we employ the \emph{domain gap} metric~\cite{yang_st3d_2021}, which relates cross-domain performance to the detector's maximum achievable performance on the target domain ($\mathrm{AP}_{T\rightarrow T}$). This relative metric provides a detached view of domain generalization ability and allows for meaningful comparisons of detectors evaluated on target datasets with varying difficulty levels. The domain gap $DG$, expressed as a percentage of the maximum achievable performance, is defined as

\begin{align}
    DG = \frac{AP_{T\rightarrow T}-AP_{S \rightarrow T}}{AP_{T\rightarrow T}} \cdot 100 
\end{align}

\subsection{Detector Architecture Evaluation} 

\begin{table}[t]
\setlength{\tabcolsep}{3pt} 
\centering
\begin{tabular}{@{}ccccc@{}}
\toprule
Detector & \begin{tabular}[c]{@{}c@{}}mAP $\uparrow$\\IOU=0.7\end{tabular} & \begin{tabular}[c]{@{}c@{}}mAP $\uparrow$\\IOU=0.4\end{tabular} & \begin{tabular}[c]{@{}c@{}}DG in \% $\downarrow$\\ IOU=0.7\end{tabular} & \begin{tabular}[c]{@{}c@{}}DG in \% $\downarrow$\\IOU=0.4\end{tabular} \\ \midrule
SECOND & 30.6 & {\ul 70.4} & 47.6 & \textbf{16.0} \\
PointPillars & 23.0 & 63.9 & 56.4 & 21.5 \\
IA-SSD & {\ul 34.7} & 66.1 & \textbf{41.5} & 18.3 \\
CenterPoint & 28.6 & 68.1 & 49.7 & 18.0 \\
PV-RCNN++ & \textbf{37.2} & \textbf{71.2} & {\ul 42.8} & {\ul 16.2} \\
DSVT & 33.4 & 68.4 & 47.7 & 20.1 \\ \bottomrule
\end{tabular}
\caption[Detector comparison evaluation]{Detector comparison results overview. We calculate the cross-domain performance by averaging over all cross-domain results. We report the average domain gap and the cross-domain performance using the AP metric at the IOU thresholds of 0.7 and 0.4 for the \emph{Vehicle} class. \vspace{-0.8cm}}
\label{tab: detector comparison results overview}
\setlength{\tabcolsep}{6pt} 
\end{table}
With the goal of finding a detector that is robust against domain changes, we examine the impact of each architectural choice on the overall performance in the cross-domain setting (see~\cref{tab: detector comparison results overview}). 
The first architectural comparison concerns the voxel and pillar discretization methods. We find that the voxel-based detector SECOND~\cite{yan_second_2018} outperforms the pillar-based PointPillars~\cite{lang_pointpillars_2019} by a substantial margin. While PointPillars is the worst-performing detector across all metrics, SECOND exhibits surprisingly good performance in the low-IOU settings. This indicates that SECOND is good at detecting objects but fails to precisely locate them in 3D space. This discrepancy stems from the quantization process. During voxelization, the exact geometric structure is lost, hampering the precise localization of objects. In terms of different data representations, we also test the point-based detector IA-SSD~\cite{zhang_not_2022}. This detector shows a very strong performance in the high-IOU setting, indicating that it is also good at predicting the 3D location of objects. This can be attributed to the detector's direct access to the point data. 

Subsequently, we test the effects of different detection heads. More specifically, we compare anchor-heads, as employed in SECOND or PointPillars, with center-heads, as introduced in CenterPoint~\cite{yin_centerpoint_2021}. Contrary to expectations, center-heads result in degraded performance compared to anchor-heads. 

Next, we apply PV-RCNN++~\cite{shi_pv-rcnn_2022} to test the impact of an additional second stage. This detector outperforms all others in terms of cross-domain performance while staying competitive in terms of domain gap. Similar to IA-SSD, the second stage of PV-RCNN++ benefits from direct access to raw point data, which likely enhances its performance. We can conclude that, for our experiments, the addition of a second stage significantly benefits the object detectors regarding generalization abilities.

Lastly, we examine the effect of Transformer-based backbones. While DSVT~\cite{wang_dsvt_2023} achieves excellent in-domain results, its cross-domain performance and domain gap metrics are only moderate. Our experiments suggest that Transformer-based backbones do not benefit object detectors in terms of domain generalization.  

These findings highlight the critical role of detector architecture in achieving robust domain generalization. Comparing the best and worst-performing detectors, we observe a performance difference of 61.7\% in the high-IOU setting and 11.4\% in the low-IOU settings. We further find that purely voxel-based detectors excel at detecting objects and the addition of point information drastically improves the localization error. The object detection architecture evaluation conducted by Eskandar \emph{et al.}~\cite{eskandar_empirical_2024} yields similar conclusion concerning the effect of point information. However, our experiments do not support their finding that Transformer-based backbones improve cross-domain generalization.

\subsection{Cross-domain Results}
\setlength{\tabcolsep}{3pt} 
\begin{table}[t]
\centering
\begin{tabular}{@{}lccc@{}}
\toprule
 & Source$\rightarrow$Target & \begin{tabular}[c]{@{}c@{}}avg. DG in \% \\ IOU=0.7 $\downarrow$\end{tabular} & \begin{tabular}[c]{@{}c@{}}avg. DG in \% \\ IOU=0.4 $\downarrow$\end{tabular} \\ \midrule
\multirow{3}{*}{\textbf{\begin{tabular}[c]{@{}l@{}}Beam \\ Density\end{tabular}}} & Dense$\rightarrow$Dense & 23.9 & 14.4 \\
 & Dense$\rightarrow$Sparse & 36.4 & 11.1 \\
 & Sparse$\rightarrow$Dense & 67.4 & 23.0 \\\bottomrule
\end{tabular}
\caption[Density- and height-induced domain gaps]{Domain gap in percent for the object detector PV-RCNN++. The different cross-domain settings are grouped and averaged by beam density. We report the average domain gap calculated with the AP metric at the IOU thresholds of 0.7 and 0.4 for the \emph{Vehicle} class. \vspace{-0.4cm}}
\label{tab: domain gap beam for beam density and mounting position}
\setlength{\tabcolsep}{6pt} 
\end{table}

Beginning our examination of the domain gap induced by beam density, we conduct a simple cross-domain evaluation. As shown in~\cref{tab: domain gap beam for beam density and mounting position}, the sparsely-trained detector (trained on Rooftop) applied on denser datasets (Sparse$\rightarrow$Dense) exhibits approximately twice the domain gap compared to applying a densely-trained detector (trained on Truck or ZOD) on a sparse dataset (Dense$\rightarrow$Sparse). This trend persists when isolating the detection error by evaluating with reduced IOU threshold. In terms of domain generalization, we could conclude from this initial analysis that for these particular datasets, it is beneficial to train a detector on the dense datasets as they generalize towards sparse and dense datasets. As the effect of beam density is just one of many factors contributing to this observed domain gap, further analysis is required to make stronger statements.  

\subsection{Density-resampling Results}
\begin{table}[t]
\centering
\begin{tabular}{@{}lccc@{}}
\toprule
 & \multicolumn{3}{c}{Target} \\ \cmidrule(l){2-4} 
 & \textbf{$\text{ZOD}_{32}$} & \textbf{$\text{ZOD}_{64}$} & \textbf{$\text{ZOD}_{128}$} \\ \cmidrule(l){2-4} 
Source & \begin{tabular}[c]{@{}c@{}}DG in \% $\downarrow$\\ IOU=0.7\end{tabular} & \begin{tabular}[c]{@{}c@{}}DG in \% $\downarrow$\\ IOU=0.7\end{tabular} & \begin{tabular}[c]{@{}c@{}}DG in \% $\downarrow$\\ IOU=0.7\end{tabular} \\ \midrule
\textbf{$\text{ZOD}_{32}$} & - & 1.8 & 5.1  \\
\textbf{$\text{ZOD}_{64}$} & 4.3 & - & 1.9 \\
\textbf{$\text{ZOD}_{128}$} & 9.4 & 0.9 & -\\ \midrule
 & \begin{tabular}[c]{@{}c@{}}DG in \% $\downarrow$\\ IOU=0.4\end{tabular} & \begin{tabular}[c]{@{}c@{}}DG in \% $\downarrow$\\ IOU=0.4\end{tabular} & \begin{tabular}[c]{@{}c@{}}DG in \% $\downarrow$\\ IOU=0.4\end{tabular} \\ \midrule
\textbf{$\text{ZOD}_{32}$} & - & 0.6 & 2.3 \\
\textbf{$\text{ZOD}_{64}$} & 2.7 & - & 1.6 \\
\textbf{$\text{ZOD}_{128}$} & 3.0 & -0.9 & - \\ \bottomrule
\end{tabular}
\caption[Domain gap for density-resampling setting caused by varying beam density]{Density-caused domain gap for the density-resampling setting. We report the domain gap calculated with the AP metric at the IOU thresholds of 0.7 (top) and 0.4 (bottom) for the \emph{Vehicle} class. \vspace{-0.4cm}}
\label{tab: in-domain gap analysis beam density}
\end{table}

\begin{table*}[ht!]
\centering
\begin{tabular}{@{}lcccccc@{}}
\toprule
 & \multicolumn{6}{c}{Target} \\ \cmidrule(l){2-7} 
 & \multicolumn{1}{l}{\textbf{$\text{ZOD}_{32}$}} & \multicolumn{1}{l}{\textbf{$\text{ZOD}_{64}$}} & \multicolumn{1}{l|}{\textbf{$\text{ZOD}_{128}$}} & \multicolumn{1}{l}{\textbf{$\text{ZOD}_{32}$}} & \multicolumn{1}{l}{\textbf{$\text{ZOD}_{64}$}} & \multicolumn{1}{l}{\textbf{$\text{ZOD}_{128}$}} \\ \cmidrule(l){2-7} 
Source & \begin{tabular}[c]{@{}c@{}}DG in \% $\downarrow$\\ IOU=0.7\end{tabular} & \begin{tabular}[c]{@{}c@{}}DG in \% $\downarrow$\\ IOU=0.7\end{tabular} & \multicolumn{1}{c|}{\begin{tabular}[c]{@{}c@{}}DG in \% $\downarrow$\\ IOU=0.7\end{tabular}} & \begin{tabular}[c]{@{}c@{}}DG in \% $\downarrow$\\ IOU=0.4\end{tabular} & \begin{tabular}[c]{@{}c@{}}DG in \% $\downarrow$\\ IOU=0.4\end{tabular} & \begin{tabular}[c]{@{}c@{}}DG in \% $\downarrow$\\ IOU=0.4\end{tabular} \\ \midrule
\textbf{$\text{Rooftop}_{32}$} & 55.0~\textcolor{red}{} & 58.0 & \multicolumn{1}{c|}{67.4} & 13.9 & 11.8 & 13.4 \\
\textbf{$\text{Truck}_{128}$} & 28.6 & 19.0 & \multicolumn{1}{c|}{16.3} & 16.9 & 8.5 & 8.2 \\ \midrule
 & \begin{tabular}[c]{@{}c@{}}AP $\uparrow$ \\ IOU=0.7\end{tabular} & \begin{tabular}[c]{@{}c@{}}AP $\uparrow$\\ IOU=0.7\end{tabular} & \multicolumn{1}{c|}{\begin{tabular}[c]{@{}c@{}}AP $\uparrow$\\ 
IOU=0.7\end{tabular}} & \begin{tabular}[c]{@{}c@{}}AP $\uparrow$\\ IOU=0.4\end{tabular} & \begin{tabular}[c]{@{}c@{}}AP $\uparrow$\\ IOU=0.4\end{tabular} & \begin{tabular}[c]{@{}c@{}}AP $\uparrow$\\ IOU=0.4\end{tabular} \\ \cmidrule(l){2-7} 
\textbf{$\text{Rooftop}_{32}$} & 25.1 & 26.7 & \multicolumn{1}{c|}{22.4} & 60.5 & 69.0 & 73.0 \\
\textbf{$\text{Truck}_{128}$} & 39.7 & 51.4 & \multicolumn{1}{c|}{57.6} & 58.4 & 71.6 & 77.4 \\ \bottomrule
\end{tabular}
\caption[Inference domain gap caused by varying beam densities]{Inference domain gap caused by varying beam densities in a cross-domain setting. We report the domain gap (top) and the cross-domain performance (bottom) using the AP metric at the IOU thresholds of 0.7 (left) and 0.4 (right) for the \emph{Vehicle} class.}
\label{tab: inference domain gap}
\end{table*}

\begin{table}[ht!]
\centering
\begin{tabular}{@{}lcccc@{}}
\toprule
 & \multicolumn{4}{c}{Target} \\ \cmidrule(l){2-5} 
 & \textbf{$\text{Rooftop}_{32}$} & \multicolumn{1}{c|}{\textbf{$\text{Truck}_{128}$}} & \textbf{$\text{Rooftop}_{32}$} & \textbf{$\text{Truck}_{128}$} \\ \cmidrule(l){2-5} 
Source & \begin{tabular}[c]{@{}c@{}}DG in \% $\downarrow$\\ IOU=0.7\end{tabular} & \multicolumn{1}{c|}{\begin{tabular}[c]{@{}c@{}}DG in \% $\downarrow$\\ IOU=0.7\end{tabular}} & \begin{tabular}[c]{@{}c@{}}DG in \% $\downarrow$\\ IOU=0.4\end{tabular} & \begin{tabular}[c]{@{}c@{}}DG in \% $\downarrow$\\ IOU=0.4\end{tabular} \\ \midrule
\textbf{$\text{ZOD}_{32}$} & 38.4 & \multicolumn{1}{c|}{41.5} 
& 13.1 & 28.0 \\
\textbf{$\text{ZOD}_{64}$} & 41.2 & \multicolumn{1}{c|}{37.2} & 11.1 & 23.7 \\
\textbf{$\text{ZOD}_{128}$} & 45.8 & \multicolumn{1}{c|}{32.5} & 9.2 & 20.5 \\ \bottomrule
\end{tabular}
\caption[Training domain gap caused by varying beam densities]{Training domain gap caused by varying beam densities in a cross-domain setting. We report the domain gap calculated with the AP metric at the IOU thresholds of 0.7 (left) and 0.4 (right) for the \emph{Vehicle} class. \vspace{-0.4cm}}
\label{tab: training domain gap}
\end{table}

We continue by isolating the beam-density-induced domain gap in our second set of experiments. To keep track of the resampled density, we call the original dense dataset $\text{ZOD}_{128}$ and the sparser variants $\text{ZOD}_{64}$ and $\text{ZOD}_{32}$, where the index represents the number of beams. As shown in ~\cref{tab: in-domain gap analysis beam density}, the density-resampling analysis contrasts with the findings from the cross-domain analysis. In the high-IOU setting, the sparse-to-dense cases (top-right of the results matrix) give better results than their dense-to-sparse counterparts (bottom-left of the results matrix). More broadly, we notice that the performance differences between all datasets are comparably small. This indicates that the detectors generalize very well towards the same dataset when solely varying the sampling. However, in real-world applications, variations in sampling is usually accompanied by other kinds of domain shift. In the following experiments we thus investigate the beam-density-induced domain shift in the presence of other kinds of domain shifts by measuring the training and inference domain gaps.


\subsection{Training and Inference Domain Gap Results}

We first examine the training domain gap in~\cref{tab: training domain gap}. Compared to the density-resampling setting (recall~\cref{tab: in-domain gap analysis beam density}), the overall domain gap level is significantly higher, as many more factors contribute besides the beam density. In the high-IOU experiments, trends are consistent with the density-resampling case: larger differences in beam density lead to larger domain gaps. However, when isolating the detection error in the low-IOU setting, a different trend emerges. Densely-trained detectors show overall better performance for the dense \emph{and} sparse target datasets, indicating that they are able to detect more objects than their sparsely-trained counterparts in the cross-domain case. This setup also allows us to quantify the impact of beam density on the overall domain gap. When evaluating on the Rooftop dataset, training on a denser 128-beam dataset \emph{reduces} the domain gap from 13.1\% to 9.2\%, reducing the domain shift by almost one-third (shown in~\cref{fig: domain gap assessment methods}).

Next, we analyze the inference domain gap in~\cref{tab: inference domain gap}. The high-IOU results (top-left of the table) show greater domain gap variability between the Rooftop and Truck datasets than within each dataset across beam densities. This observation supports our earlier assumption about persistent dataset-caused domain gaps in cross-domain evaluation settings (recall~\cref{sec:approach}). Overall, the results indicate that the detector is relatively robust to beam density, provided the number of beams does not change drastically. In the high-IOU experiments, the domain gap increases by only 3\% or less when doubling or halving beam density.

The previous analyses were exclusively done through the lens of the \emph{domain gap} metric. For the inference domain gap, it is also interesting to examine the performance values themselves. Especially for the low-IOU setting, we can see in~\cref{tab: inference domain gap} that the performance (measured in AP) increases steadily with an increasing number of beams, despite the domain gap staying similar. This indicates that the observed performance gain is caused by easing the detection problem in contrast to better generalizability of the detectors. As the density increases, more LiDAR rays hit objects which makes it easier for the object to be detected. 

In summary, the results provide a comprehensive view of the domain gap caused by varying beam densities. When isolating the effect of varying beam densities (see~\cref{tab: in-domain gap analysis beam density}), the domain gap appears minor, favoring sparsely-trained detectors for domain generalization. This aligns with findings from related studies~\cite{eskandar_empirical_2024, fang_lidar-cs_2024, richter_understanding_2022}. However, when analyzing the training domain gap in conjunction with other domain shifts, we find that densely-trained detectors exhibit better domain generalization in terms of detecting objects (see~\cref{tab: training domain gap}). Regarding inference domain gaps (see~\cref{tab: inference domain gap}), results show that detectors generalize well as long as beam density changes are modest. Nonetheless, denser sampling reduces detection difficulty, leading to better performance irrespective of the detector's generalizability.

\section{Conclusion}
\label{sec:conclusion}
This study presented an investigation of the impact of beam density on LiDAR object detection performance in cross-domain scenarios during which we also explored optimal object detector architectures to address domain variability effectively. Our object detector architecture evaluation revealed that combining voxel- and point-based approaches delivers superior cross-domain performance by leveraging the complementary strengths of these representations. While Transformer-based backbones demonstrated strong performance in in-domain tasks, their cross-domain benefits were limited under the conditions tested. Our findings emphasize the importance of selecting a robust detector architecture as a prior step to domain adaptation.

We further investigated the impact of beam density on LiDAR object detection performance in cross-domain scenarios, offering insights into both training and inference domain gaps. We found that detectors trained on dense datasets generalize better across domains, particularly for detecting objects, where detection error (rather than localization error) is the primary concern. During inference, detectors showed robustness against moderate beam density changes, with denser configurations improving performance by reducing the difficulty of the detection task rather than enhancing generalizability.

A key insight from this study is that domain gaps, including those caused by beam density, should not be analyzed in isolation. Instead, we advocate for a holistic approach to domain adaptation, beginning with the selection of a detector intrinsically robust to domain changes. This minimizes the initial domain gap and allows adaptation efforts to focus on more complex types of domain shifts.

\section{Acknowledgements}
The financial support by the Austrian Federal Ministry for Digital and Economic Affairs, the National Foundation for Research, Technology and Development and the Christian Doppler Research Association is gratefully acknowledged.


{
    \small
    \bibliographystyle{ieeenat_fullname}
    \bibliography{references}
}
\appendix
\section{Dataset Introduction}
\begin{figure}[h]
\centering
  \includegraphics[width=0.45\textwidth]{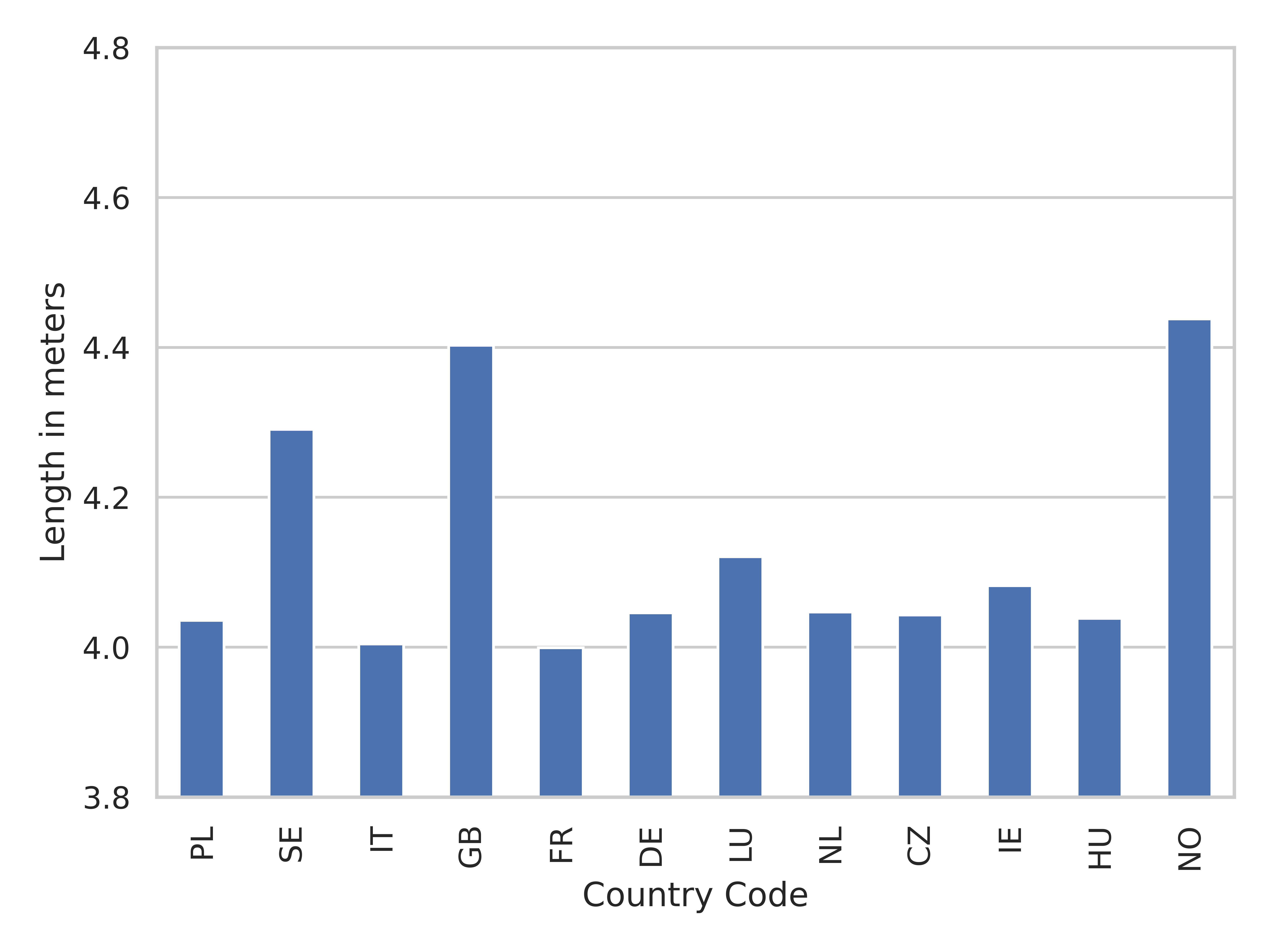}
\caption[Average vehicle length differences on Zenseact dataset]{Average length of vehicles for different countries in the ZOD.}
\label{fig: zod vehicle length per country}
\end{figure}

\begin{figure}[h]
\centering
  \includegraphics[width=0.45\textwidth]{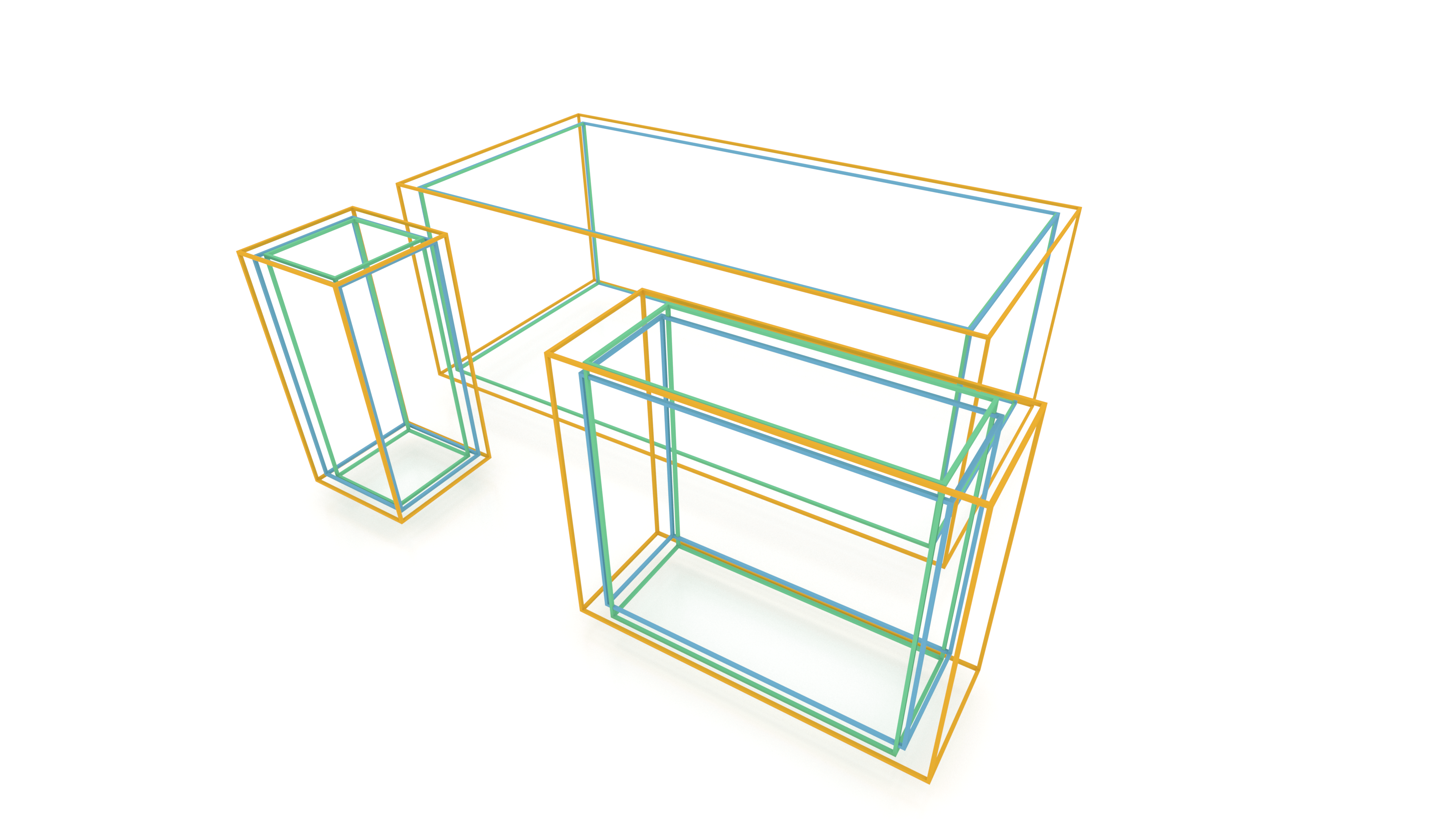}
\caption[Comparison of average object sizes]{Comparison of average object sizes for the classes \emph{Car}, \emph{Pedestrian} and \emph{Cyclist} for {ZOD} (green), Truck dataset (blue) the Rooftop dataset (orange). Object sizes of the Rooftop dataset are significantly larger on average.}
\label{fig: object size comparison}
\end{figure}

\begin{figure}
\centering
  \includegraphics[width=0.45\textwidth]{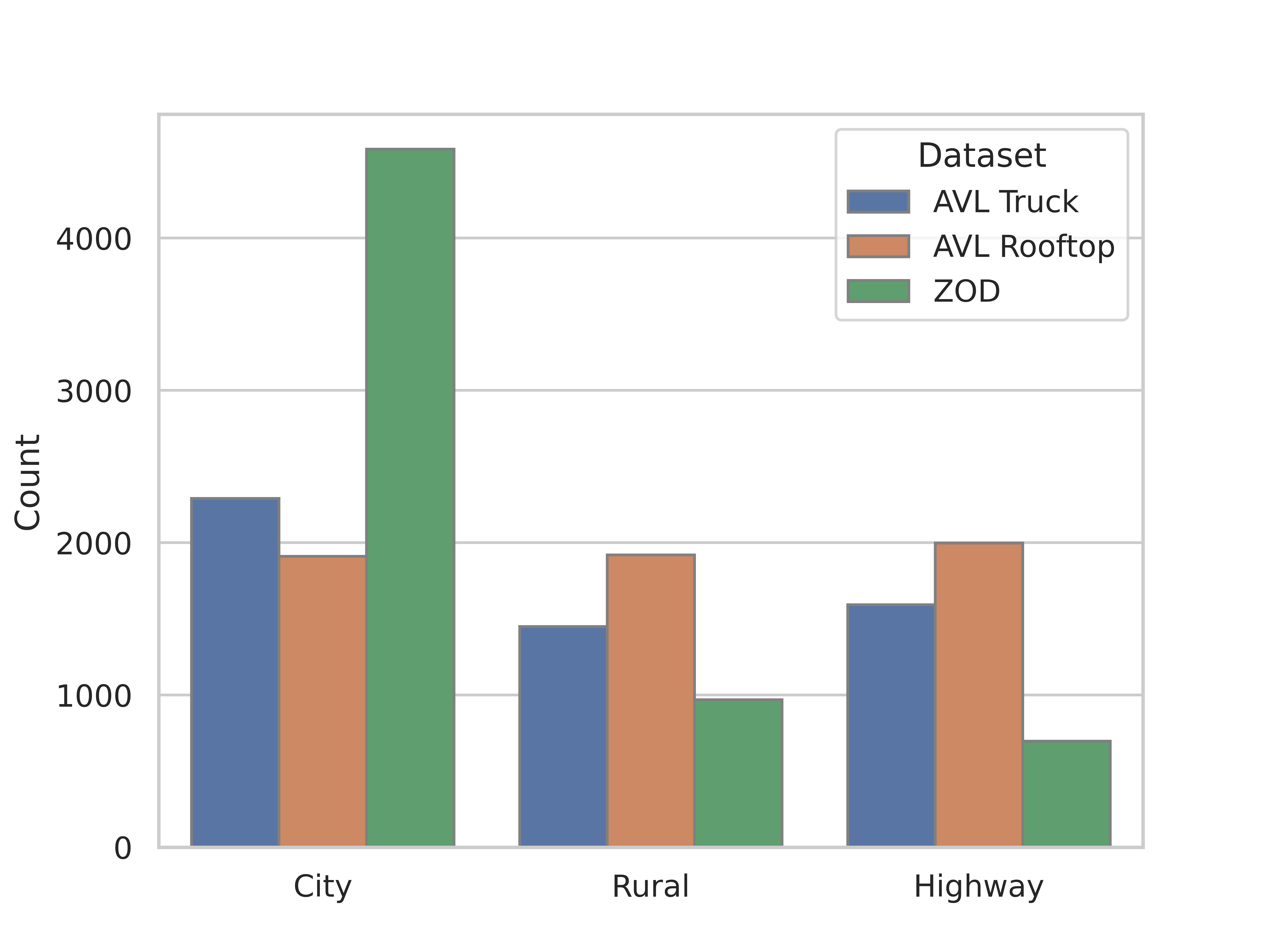}
\caption{Recording area statistics.}
\label{fig: area types}
\end{figure}

\begin{figure}[h]
  \includegraphics[width=0.45\textwidth]{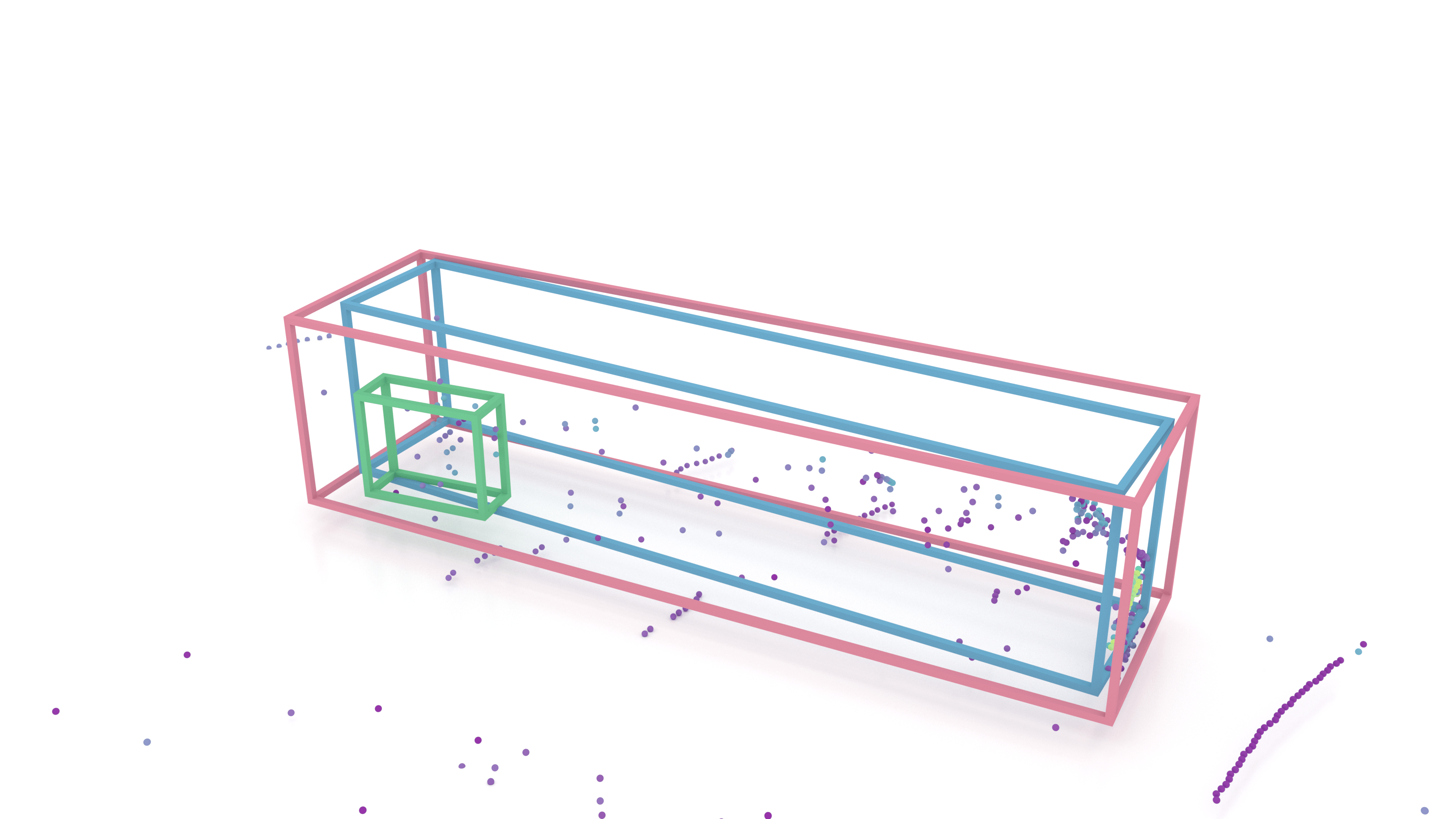}
\caption[Example of a misclassification of ambiguous object]{Example of a misclassification of an ambiguous object on a highway. The detector trained on the {ZOD} (green bounding box) is more likely to assign the class \emph{Cyclist} to the ambiguous object compared to the detector trained on the Rooftop dataset (blue bounding box), which correctly identifies the object as \emph{Truck} (red bounding box).}
\label{fig: misclassificaiton cyclist truck}
\end{figure}

\begin{figure}
\centering
  \includegraphics[width=0.45\textwidth]{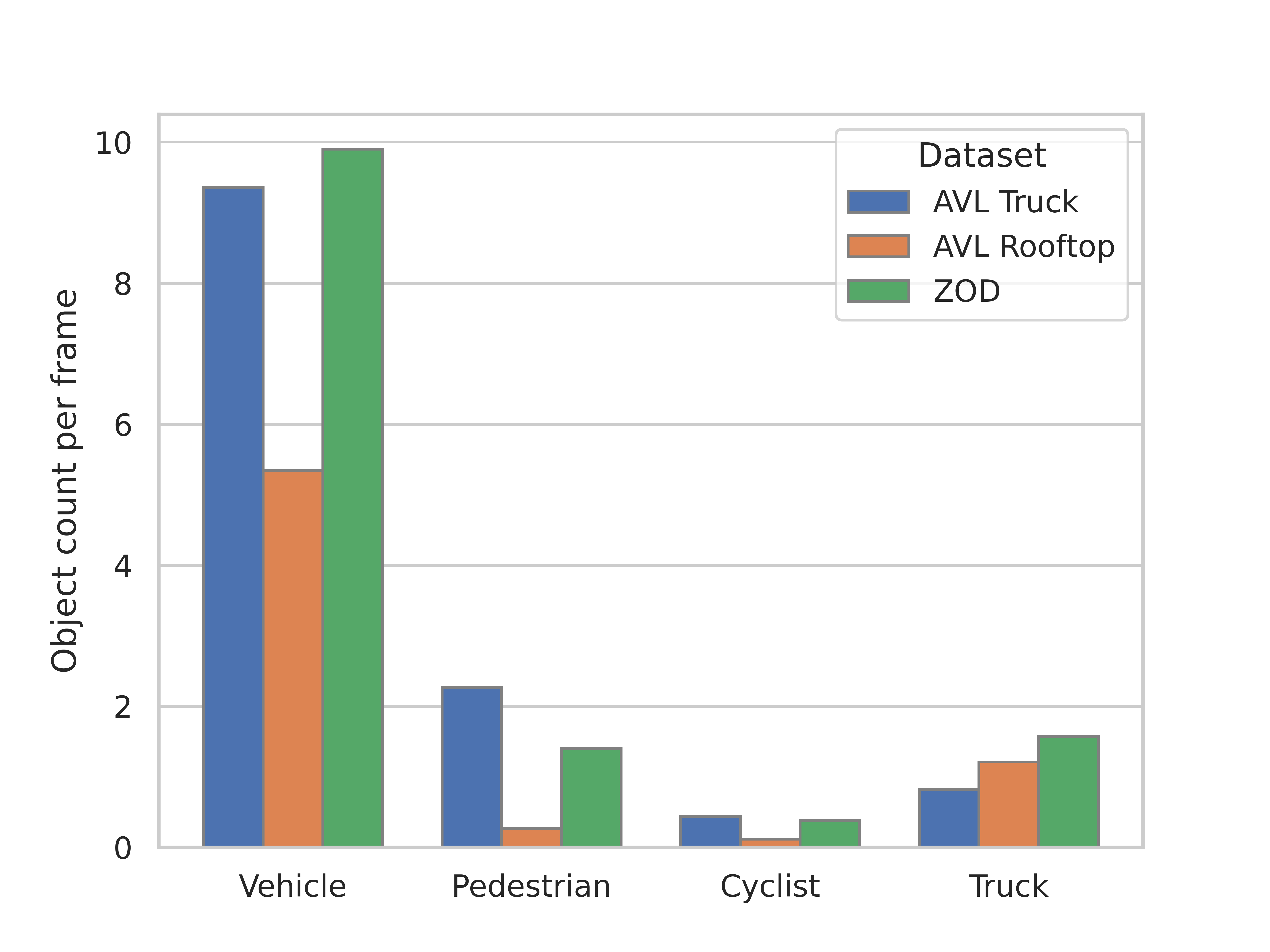}
\caption{Class statistics.}
\label{fig: object count per frame and class}
\end{figure}

\begin{figure*}
\hspace{-1.5cm}
\centering
\begin{minipage}[t]{.45\textwidth}
\centering
  \includegraphics[width=1.2\textwidth]{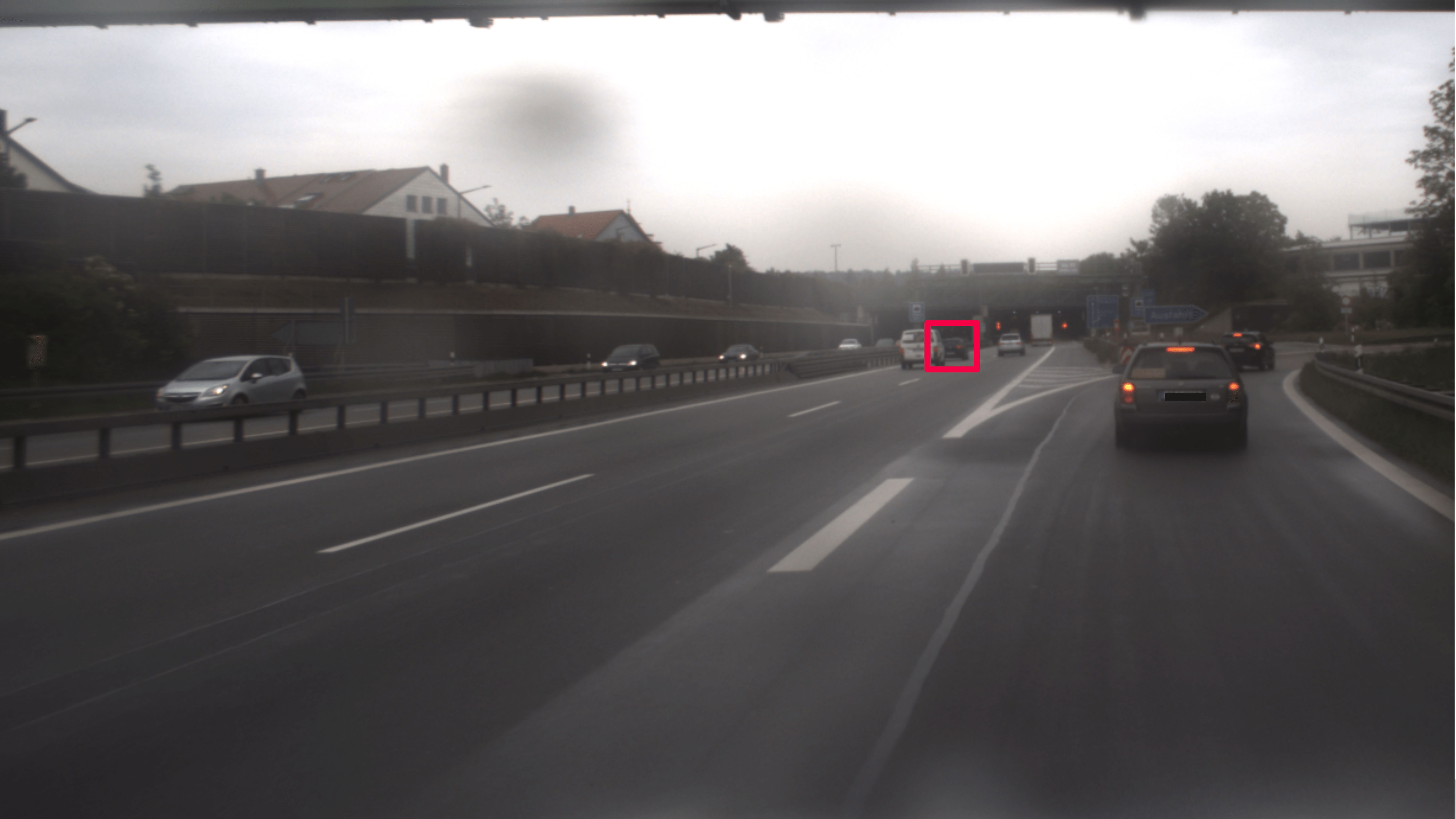}
\end{minipage}
\hspace{1cm}
\begin{minipage}[t]{.35\textwidth}
\centering
  \includegraphics[trim={0 0 8cm 0},width=0.35\textwidth]{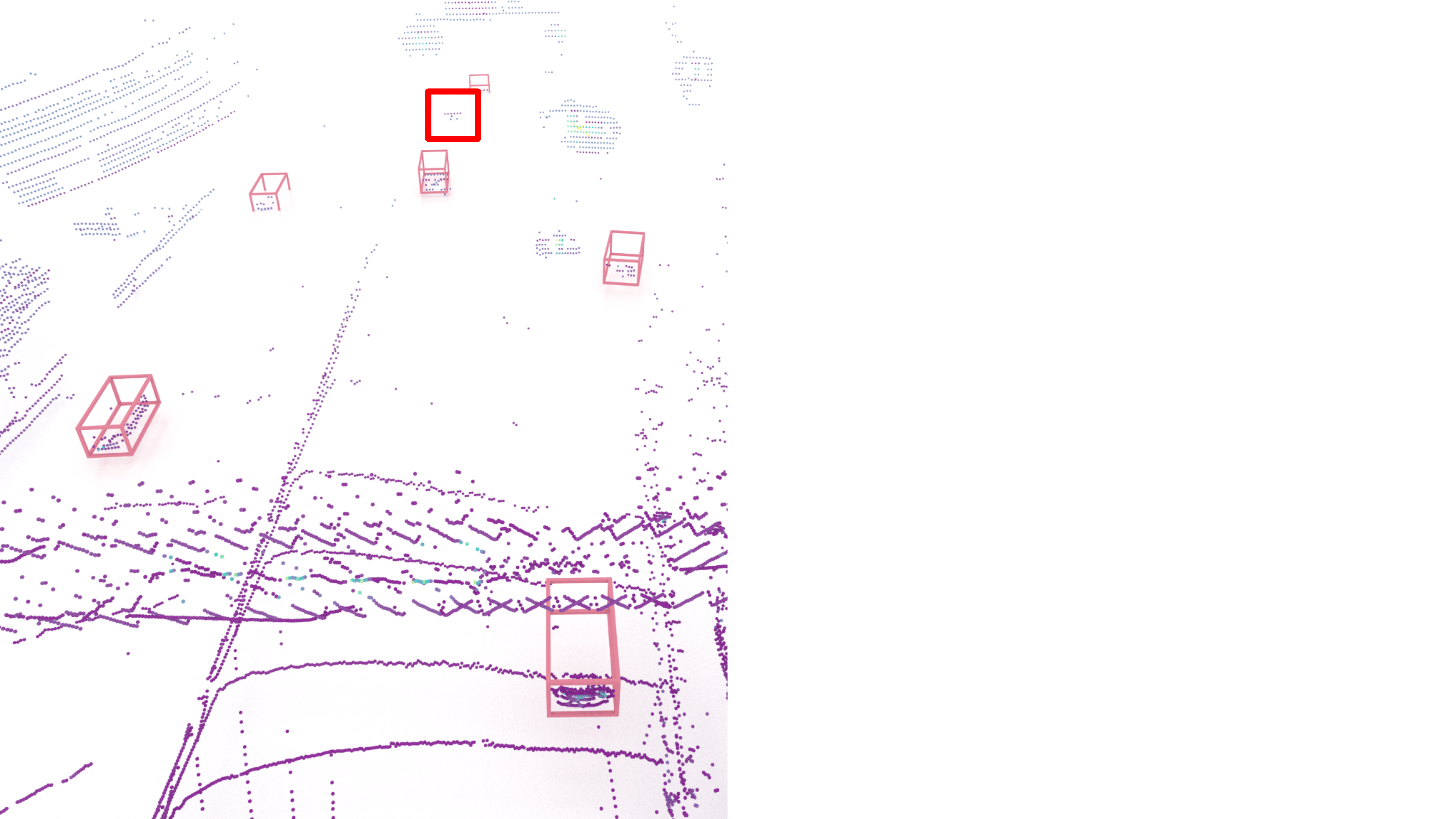}
\end{minipage}%
\caption[Example of missing ground truth label]{Example of missing ground truth label for the Rooftop dataset. The 3D bounding box of the red boxed car is missing even though it is clearly visible in the image and {LiDAR} data.\vspace{-0.4cm}}
\label{fig: missing label}
\end{figure*}

In the following, we provide the detailed label map (see~\cref{tab: label-space mapping}) from Chap. 3.1 and elaborate some of the domain shifts mentioned in Chap. 3.2 in more detail. We begin by showcasing the effect of geographically diverse locations in~\cref{fig: zod vehicle length per country}. The observed country-level size bias combined with a dataset-specific size-bias results in different average object sizes between the datasets (see~\cref{fig: object size comparison}). \cref{fig: area types} shows differences in recording locations. We can see that the ZOD contains significantly more \emph{City} frames compared to the other two datasets. As a consequence, a detector trained on the ZOD is more likely to assign objects that typically associated with \emph{City} frames, such as \emph{Cyclists}, to ambiguous objects than detectors trained on the other datasets. An example of this phenomenon is depicted in~\cref{fig: misclassificaiton cyclist truck}, where the detector trained on the ZOD detects a \emph{Cyclist}, while the other detectors correctly detect a \emph{Truck}. The bias introduced by the frame selection procedure can be seen in~\cref{fig: object count per frame and class}. The Rooftop dataset contains, on average, less objects per frame than the other two datasets. This difference is especially severe for the classes \emph{Pedestrian} and \emph{Cyclist}. Finally, we give an example for imperfect labeling of the Rooftop dataset in~\cref{fig: missing label}. The camera image shows a black car, which is captured by 8 points in the LiDAR image. However, no bounding box is assigned in the LiDAR frame. 
\begin{table*}[]
\centering
\small
\begin{tabular}{@{}l|ccr@{}}
\toprule
\textbf{\begin{tabular}[c]{@{}l@{}}Detector \\ Label-Space\end{tabular}} & \multicolumn{3}{c}{\textbf{\begin{tabular}[c]{@{}c@{}}Dataset \\ Label-Space\end{tabular}}} \\ \midrule
 & Truck & Rooftop & Zenseact Open Dataset \\ \midrule
\multirow{2}{*}{Vehicle} & Vehicle\_Drivable\_Car & Vehicle\_Drivable\_Car & Vehicle\_Car \\
 & Vehicle\_Drivable\_Van & Vehicle\_Drivable\_Van & Vehicle\_Van \\ \midrule
\multirow{6}{*}{Truck} & LargeVehicle\_Bus & LargeVehicle\_Bus & Vehicle\_Bus \\
 & LargeVehicle\_TruckCab & LargeVehicle\_TruckCab &  \\
 & Trailer & Trailer & Vehicle\_Trailer \\
 & \multirow{3}{*}{LargeVehicle\_Truck} & \multirow{3}{*}{LargeVehicle\_Truck} & Vehicle\_Truck \\
 &  &  & Vehicle\_TramTrain \\
 &  &  & Vehicle\_HeavyEquip \\ \midrule
\multirow{2}{*}{Cyclist} & Vehicle\_Ridable\_Motorcycle & Vehicle\_Ridable\_Motorcycle & VulnerableVehicle\_Motorcycle \\
 & Vehicle\_Ridable\_Bicycle & Vehicle\_Ridable\_Bicycle & VulnerableVehicle\_Bicycle \\ \midrule
Pedestrian & Human & Human & Pedestrian \\ \midrule
\multirow{4}{*}{DontCare} & Dont\_Care & PPObject &  \\
 & Other & PPObject\_Stroller &  \\
 &  & PPObject\_BikeTrailer &  \\
 &  & Vehicle\_PMD & \\
 \bottomrule
\end{tabular}
\caption[Label-space mapping between detector and the dataset label-spaces]{Label-space mapping between the detector label-space and the dataset label-spaces.\vspace{-0.4cm}}
\label{tab: label-space mapping}
\end{table*}

\section{Experiments}
\subsection{Implementation Details}
In the following, we summarize some implementation details which are shared across the object detection models.

\paragraph{Codebase.}
All models were implemented in the codebase 3DTrans\footnote{\href{https://github.com/PJLab-ADG/3DTrans}{https://github.com/PJLab-ADG/3DTrans}}, which is an extension of the open-source 3D object detecton codebase OpenPCDet~\cite{openpcdet_development_team_openpcdet_2020}. Models developed in OpenPCDet can seamlessly be integrated into 3DTrans. All the models were already implemented in 3DTrans for the Waymo Open Dataset~\cite{sun_scalability_2020}, with the exception of DSVT, which had to be adopted from the official OpenPCDet codebase. The {ZOD} dataloader has been implemented based the provided developement kit\footnote{\href{https://github.com/zenseact/zod}{https://github.com/zenseact/zod}}. The dataloaders for the Rooftop and Truck datasets were implemented from scratch. Our implementations are based on PyTorch 2.1 and SpConv~\cite{spconv_contributors_spconv_2022} version 2.3.6 for CUDA 12.0.

\paragraph{Hardware.}
We conducted the development and testing of the models on a workstation featuring a single RTX 4090 {GPU}. We trained the final models on a {GPU} with four RTX A6000 GPUs. 

\paragraph{Schedule and Optimization.}
We train all object detectors on each dataset for 100 epochs. All the models employ the ADAM optimizer~\cite{kingma_adam_2015} and use a OneCycle learning-rate scheduler~\cite{smith_super-convergence_2018} with varying learning rate, momentum and weight-decay parameters depending on the model. 

\paragraph{Data Representation.}
The voxel-based methods SECOND, CenterPoint, PV-RCNN++, and  DSVT require a discretization of the point cloud into a voxel-representation before the object detection models can be applied. To this end, we adapt a voxel size of (0.1m, 0.1m, 0.15m) following the implementation of PV-RCNN++. For the pillar-based method PointPillars, we use a pillar-size of (0.32m, 0.32m, 6.0m).

\subsection{Detector Architecture Search}
In~\cref{tab: detector comparison raw}, we provide the raw data used to calculate the averaged results for the domain gap and performance, which we base our detector architecture selection on. These results are also used to conduct the initial cross-domain experiment in Sec. 5.3.  
\begin{table}[t]
\centering
\small
\begin{tabular}{@{}lcccc@{}}
\toprule
 & \multicolumn{1}{l}{} & \multicolumn{3}{c}{Target} \\ \cmidrule(l){3-5} 
 & \multicolumn{1}{l}{} & \textbf{Truck} & \textbf{Rooftop} & \textbf{\begin{tabular}[c]{@{}c@{}}Zenseact \\ Open Dataset\end{tabular}} \\ \cmidrule(l){3-5} 
Source & Detector & \begin{tabular}[c]{@{}c@{}c@{}}AP $\uparrow$\\IOU\\0.7/0.4\end{tabular} & \begin{tabular}[c]{@{}c@{}c@{}}AP $\uparrow$\\IOU\\0.7/0.4\end{tabular} & \begin{tabular}[c]{@{}c@{}c@{}}AP $\uparrow$
\\IOU\\0.7/0.4\end{tabular} \\ \midrule
\multirow{6}{*}{\textbf{\begin{tabular}[c]{@{}l@{}}Truck\end{tabular}}} 
 & SECOND & 55.5/84.9 & 39.2/73.6 & 46.9/74.7 \\
 & PointPillars & 49.7/82.4 & 32.7/67.3  & 33.5/67.7 \\
 & IA-SSD & 58.6/82.4 & 41.3/69.5 & 53.9/72.1 \\
 & CenterPoint & 54.5/82.4 & 36.9/75.7 & 43.6/74.2 \\
 & PV-RCNN++ & 65.5/86.4 & 45.5/74.0 & 57.6/77.4 \\
 & DSVT & 60.4/86.3 & 41.0/72.1 & 53.2/77.3 \\ \midrule
\multirow{6}{*}{\textbf{\begin{tabular}[c]{@{}l@{}}Rooftop\end{tabular}}}
 & SECOND & 16.7/60.5 & 58.3/84.6 & 16.9/66.9 \\
 & PointPillars & 13.6/56.3 & 51.8/82.3 & 13.6/65.1 \\
 & IA-SSD & 20.6/59.9 & 55.1/82.2 & 29.9/71.7 \\
 & CenterPoint & 13.9/54.4 & 58.3/84.0 & 11.7/58.9 \\
 & PV-RCNN++ & 21.0/58.3 & 61.4/84.3 & 22.4/73.0 \\
 & DSVT & 20.9/55.2 & 64.6/86.4 & 19.9/62.6 \\ \midrule
\multirow{6}{*}{\textbf{\begin{tabular}[c]{@{}l@{}}Zenseact\\ Open\\ Dataset\end{tabular}}} 
 & SECOND & 36.7/69.7 & 27.0/76.8 & 61.2/82.0 \\
 & PointPillars & 25.3/59.5 & 19.1/67.6 & 56.7/79.8 \\
 & IA-SSD & 33.7/54.0 & 28.7/69.3 & 63.8/78.6 \\
 & CenterPoint & 38.7/69.8 & 26.9/75.5 & 57.7/82.7 \\
 & PV-RCNN++ & 44.2/68.6 & 32.7/75.8 & 68.7/84.3 \\
 & DSVT & 38.1/66.2 & 27.6/77.0 & 66.4/84.3 \\ \bottomrule
\end{tabular}
\caption[Detector comparison in terms of the cross-domain performance]{Detector comparison in terms of the cross-domain performance. We report the performance using the AP metric at an IOU threshold of 0.7/0.4 for the \emph{Vehicle} class.}
\label{tab: detector comparison raw}
\end{table}

\subsection{Domain Gap Results}
In~\cref{tab: in-domain gap analysis beam density performance} and~\cref{tab: training domain gap performance} we provide the performance values in Average Precision based on which the domain gaps in Sec. 5.4 and 5.5 are calculated.  
\begin{table}[h]
\centering
\small
\begin{tabular}{@{}lccc@{}}
\toprule
 & \multicolumn{3}{c}{Target} \\ \cmidrule(l){2-4} 
 & \textbf{$\text{ZOD}_{32}$} & \textbf{$\text{ZOD}_{64}$} & \textbf{$\text{ZOD}_{128}$} \\ \cmidrule(l){2-4} 
Source & \begin{tabular}[c]{@{}c@{}}AP $\uparrow$\\ IOU=0.7\end{tabular} & \begin{tabular}[c]{@{}c@{}}AP $\uparrow$  \\ IOU=0.7\end{tabular} & \begin{tabular}[c]{@{}c@{}}AP $\uparrow$\\ IOU=0.7\end{tabular} \\ \midrule
\textbf{$\text{ZOD}_{32}$}  & 55.7 & 62.3 & 65.3  \\
\textbf{$\text{ZOD}_{64}$}  & 53.3 & 63.5 & 67.4 \\
\textbf{$\text{ZOD}_{128}$} & 50.4 & 62.9 & 68.7 \\ \midrule
 & \begin{tabular}[c]{@{}c@{}}AP $\uparrow$\\ IOU=0.4\end{tabular} & \begin{tabular}[c]{@{}c@{}}AP $\uparrow$\\ IOU=0.4\end{tabular} & \begin{tabular}[c]{@{}c@{}}AP $\uparrow$\\ IOU=0.4\end{tabular} \\ \midrule
\textbf{$\text{ZOD}_{32}$}  & 70.3 & 77.8 & 82.3 \\
\textbf{$\text{ZOD}_{64}$}  & 68.4 & 78.3 & 82.9 \\
\textbf{$\text{ZOD}_{128}$} & 68.2 & 79.0 & 84.3 \\ \bottomrule
\end{tabular}
\caption[Domain gap for density-resampling setting caused by varying beam density]{Density-caused domain gap for the density-resampling setting. We report the performance with the AP metric at the IOU thresholds of 0.7 (top) and 0.4 (bottom) for the \emph{Vehicle} class. \vspace{-0.4cm}}
\label{tab: in-domain gap analysis beam density performance}
\end{table}

\begin{table}[h]
\centering
\small
\begin{tabular}{@{}lcccc@{}}
\toprule
 & \multicolumn{4}{c}{Target} \\ \cmidrule(l){2-5} 
 & \textbf{$\text{Rooftop}_{32}$} & \multicolumn{1}{c|}{\textbf{$\text{Truck}_{128}$}} & \textbf{$\text{Rooftop}_{32}$} & \textbf{$\text{Truck}_{128}$} \\ \cmidrule(l){2-5} 
Source & \begin{tabular}[c]{@{}c@{}}AP $\uparrow$\\ IOU=0.7\end{tabular} & \multicolumn{1}{c|}{\begin{tabular}[c]{@{}c@{}}AP $\uparrow$\\ IOU=0.7\end{tabular}} & \begin{tabular}[c]{@{}c@{}}AP $\uparrow$\\ IOU=0.4\end{tabular} & \begin{tabular}[c]{@{}c@{}}AP $\uparrow$\\ IOU=0.4\end{tabular} \\ \midrule
\textbf{$\text{ZOD}_{32}$} & 37.9 & \multicolumn{1}{c|}{38.3} & 73.3 & 62.2 \\
\textbf{$\text{ZOD}_{64}$} & 36.1 & \multicolumn{1}{c|}{41.1} & 74.9 & 65.9 \\
\textbf{$\text{ZOD}_{128}$} & 33.3 & \multicolumn{1}{c|}{44.2} & 76.6 & 68.7 \\ \bottomrule
\end{tabular}
\caption[Training domain gap caused by varying beam densities]{Training domain gap caused by varying beam densities in a cross-domain setting. We report cross-domain performance with the AP metric at the IOU thresholds of 0.7 (left) and 0.4 (right) for the \emph{Vehicle} class. \vspace{-0.4cm}}
\label{tab: training domain gap performance}
\end{table}

  \newpage

\end{document}